\documentclass[10pt,twocolumn,letterpaper]{article}

\usepackage{cvpr}
\usepackage{times}
\usepackage{epsfig}
\usepackage{graphicx}
\usepackage{amsmath}
\usepackage{amssymb}

\usepackage{booktabs}
\usepackage{colortbl}
\usepackage{xcolor}
\usepackage{xspace}
\usepackage{threeparttable}
\usepackage{multirow}
\usepackage{makecell}
\usepackage{graphicx}
\usepackage{amssymb,amsmath}
\usepackage{subfig}
\usepackage[shortlabels]{enumitem}
\usepackage[acronym]{glossaries}
\usepackage[ruled,vlined]{algorithm2e}
\usepackage{mathtools}
\usepackage{dsfont}

\newacronym{ssl}{SSL}{Semi-Supervised Learning}
\newacronym{emprm}{ERM}{Empirical Risk Minimization}
\newacronym{gan}{GAN}{Generative Adversarial Network}
\newacronym{mmd}{MMD}{Maximum Mean Discrepancy}
\newacronym{cnn}{CNN}{Convolutional Neural Network}
\newacronym{tsne}{t-SNE}{t-Distributed Stochastic Neighbor Embedding}
\newacronym{vat}{VAT}{Virtual Adversarial Training}
\newacronym{uda}{UDA}{Unsupervised Domain Adaptation}

\newcommand{\E}{\mathbb{E}}

\newcommand{\abs}[1]{\left|#1\right|}

\newcommand*{\brackets}[1]{\left(#1\right)}
\newcommand*{\bracketss}[1]{\left[#1\right]}
\newcommand{\krbf}{k_{\mathrm{rbf}}}

\newcommand*{\dataset}[1]{\mathcal{#1}}

\newcommand*{\DX}{\mathcal{X}}
\newcommand*{\DY}{\mathcal{Y}}

\newcommand*{\lse}{\mathrm{LSE}}
\newcommand*{\softmax}{\mathrm{softmax}}
\newcommand*{\loss}{\mathcal{L}}

\renewcommand*{\vec}[1]{#1}
\newcommand*{\img}{x}
\newcommand*{\lab}{\vec{y}}
\newcommand*{\z}{\vec{z}}

\newcommand*{\W}{\mathbf{W}}
\newcommand*{\bias}{\mathbf{b}}

\makeatletter
\DeclareRobustCommand\onedot{\futurelet\@let@token\@onedot}
\def\@onedot{\ifx\@let@token.\else.\null\fi\xspace}

\def\eg{\emph{e.g}\onedot} 
\def\ie{\emph{i.e}\onedot}

\def\etal{\emph{et al}\onedot}
\makeatother

\newtheorem{theorem}{Theorem}

\usepackage[pagebackref=true,breaklinks=true,letterpaper=true,colorlinks,bookmarks=false]{hyperref}

\cvprfinalcopy

\begin{document}

\title{Adversarial Feature Distribution Alignment for Semi-Supervised Learning}

\author{Christoph Mayer \hspace{1cm} Matthieu Paul \hspace{1cm} Radu Timofte\\
Computer Vision Lab, ETH Z{\"u}rich, Switzerland\\
{\tt\small \{chmayer,paulma,timofter\}@vision.ee.ethz.ch}
}

\maketitle

\begin{abstract}
Training deep neural networks with only a few labeled samples can lead to overfitting. This is problematic in \gls{ssl} where only a few labeled samples are available. In this paper, we show that a consequence of overfitting in \gls{ssl} is feature distribution misalignment between labeled and unlabeled samples. Hence, we propose two new feature distribution alignment methods to reduce overfitting. Our methods are particularly effective when using only a small amount of labeled samples. Furthermore, we add consistency regularization to our adversarial alignment method and demonstrate that we always outperform pure consistency regularization and achieve particularly high improvements when using only a small amount of labeled samples. We test our method on CIFAR-10 and SVHN. On SVHN we achieve a test error of 3.88\% (250 labeled samples) and 3.39\% (1000 labeled samples) which is close to the fully supervised model 2.89\% (73k labeled samples). In comparison, the current state-of-the-art achieves only 4.29\% and 3.74\%.
\end{abstract}

\section{Introduction}

Recently, deep learning gained in popularity for achieving performances beyond human capabilities~\cite{he2016resnet,silver2017alphago}. One of the main ingredients for the success of deep learning are large-scale data sets with thousands of annotated images. In practice, it is costly to assemble such massive data collections despite their crucial importance. Whereas acquiring images is often relatively cheap, annotating is cumbersome, error-prone and expensive~\cite{bearman2016point}. Several research fields have therefore emerged, striving for fully-supervised performance when annotating only a small fraction of the data set. These methods either actively choose which images should be labeled~\cite{sener2018} or leverage the information of unlabeled images~\cite{oliver2018realistic} to improve the performance.

\begin{figure}[t]
\centering 
\includegraphics[width=\columnwidth,keepaspectratio]{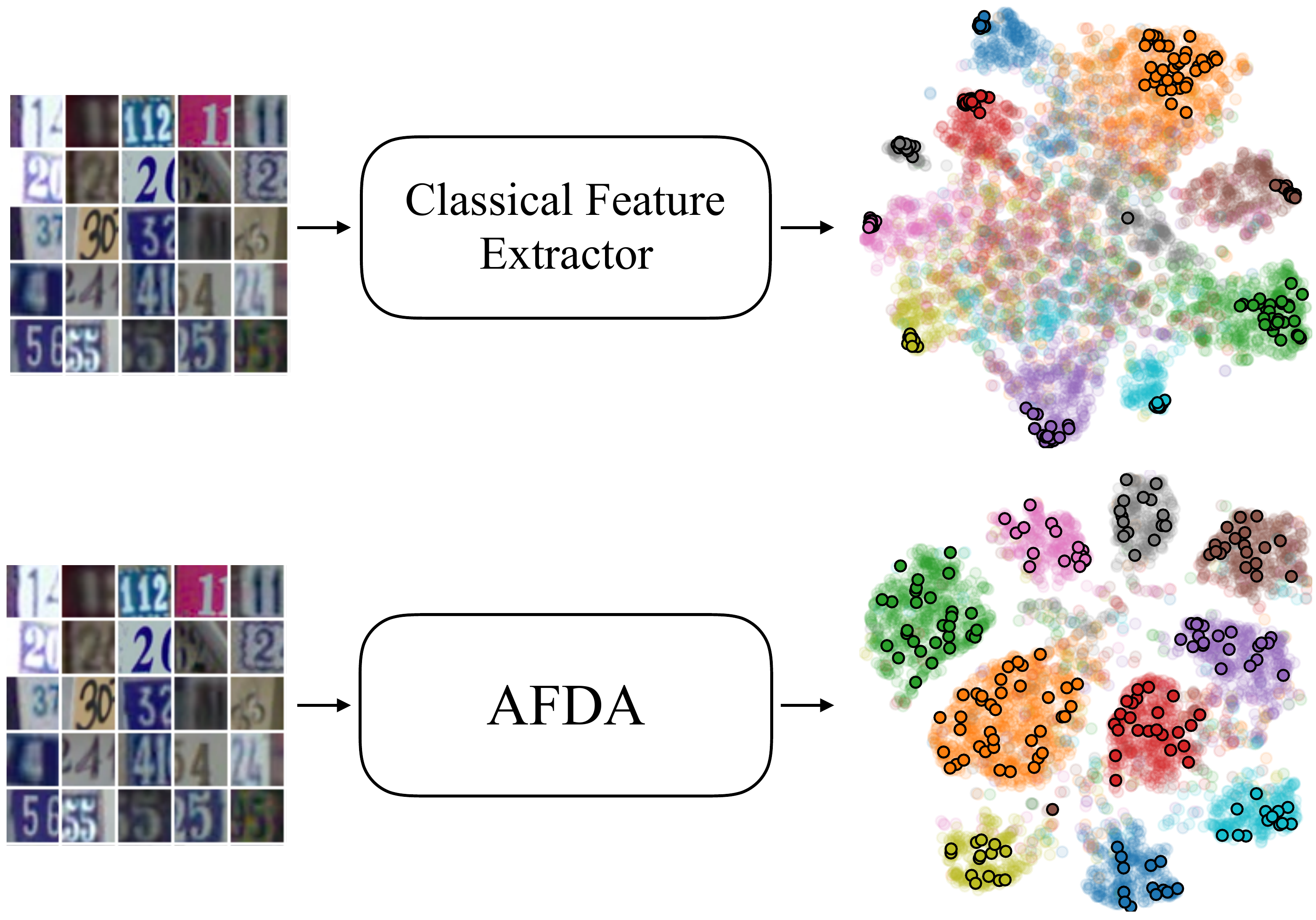}
\caption{Visualization of the feature space embedding produced by a classical feature extractor and our proposed feature extractor with Adversarial Feature Distribution Alignment (AFDA). Our feature extractor reduces overfitting on the labeled data while full filling both the low-density and the cluster assumption. The feature representations are projected using t-SNE~\cite{maaten2008visualizing}.}\label{fig:teaser}
\end{figure}

In order to achieve small test errors, deep learning models require an adequate number of parameters~\cite{zhang2018mixup}, and training such models with only a small set of labeled images increases the risk of overfitting. 
Optimally, training a deep learning model would involve minimizing its error over the true data distribution. Unfortunately, the true data distribution is usually unknown and precludes expected risk minimization. Hence, empirical risk minimization~\cite{vapnik2013nature} reduces the training error instead, but might lead to poor generalization. Typical strategies to increase generalization or to reduce overfitting are regularization techniques such as: early stopping, drop-out~\cite{srivastava2014dropout}, weight-decay or synthetically increasing the data set using data augmentation~\cite{zhang2018mixup}. In contrast, \gls{ssl} methods aim at improving the generalization of machine learning models by using large sets of unlabeled data in addition to a small set of labeled data.
Typically, \gls{ssl} methods rest on the \textit{low density separation} and \textit{cluster assumption}~\cite{chapelle2009ssl}. They state that there exists a feature space where neighbouring samples are more likely to belong to the same class category than distant samples, and that class transitions occur more likely in low density regions.

\gls{ssl} methods were shown to largely outperform the models trained solely on the labeled data~\cite{oliver2018realistic}. Nonetheless, there is still a gap between the performance of the  fully-supervised model where all samples were labeled and used for training and a semi-supervised model where only a few samples of the data set were labeled. In such extrem settings overfitting of the classifiers is still problematic. Fig.~\ref{fig:teaser} shows our proposed method and a traditional method. The 2D embedding of the high dimensional feature space reveals that traditional methods produce features that group only the labeled samples from the same category close together (circles with black boundary) but fail to generalize this property to a majority of unlabeled samples (circles without boundary). Thus, although both the labeled and unlabeled samples are sampled \textit{i.i.d.} from the same data distribution their corresponding distributions in feature space are misaligned. Thus, using randomly initialized weights for the feature extractor would result in aligned feature distributions as the samples are \textit{i.i.d.} but training with only a few labeled samples leads to misaligned feature distributions. 

Thus, the main observations that motivate this paper are that \gls{ssl} methods suffer from feature distribution misalignment and that it can be cured by actively aligning the feature distributions of labeled and unlabeled samples.

A line of research that aims at aligning different feature distributions in order to reduce the test error is \gls{uda}~\cite{ben2010theory}. The goal of \gls{uda} is training a classifier on a fully labeled data set sampled from the source domain in order to achieve a small test error on a different target domain~\cite{ben2010theory}. In \gls{uda}, only unlabeled samples from the target domain are accessible to the classifier. As a consequence, there exist a few but important differences between \gls{uda} and \gls{ssl} methods: (i) the distribution mismatch in \gls{ssl} is introduced by training the feature extractor only on the labeled samples such that it fails to generalize to the unlabeled samples, while the mismatch in \gls{uda} exists by definition; (ii) in \gls{uda} the full source data set is typically labeled such that thousands of annotations are available~\cite{ganin15domain}. Conversely, in \gls{ssl} only a very small amount of images are labeled which complicates training a deep learning model considerably~\cite{oliver2018realistic}; (iii) the threat of overfitting is less problematic in \gls{uda} than in \gls{ssl}; (iv) in \gls{uda} the domain of the training and testing set vary whereas in \gls{ssl} both are identical. 
Thus, our contributions are as follows

\textbf{(i)} We identify the problem of feature distribution misalignment in \gls{ssl}. Then, we identify the similarities and differences between \gls{ssl} and \gls{uda} in order to propose a new method that aligns the feature distributions within deep learning methods and that is easy to implement in existing models.
\textbf{(ii)} We demonstrate on the two most relevant data sets for \gls{ssl} (SVHN and CIFAR-10) that our method achieves the SOTA, and is especially effective when using only a few labeled samples compared to other \gls{ssl} methods. 
\textbf{(iii)} We observe that using both consistency regularization and adversarial alignment at the same time leads to mutual benefits compared to using only one. \textbf{(iv)} Furthermore, we demonstrate that adversarial alignment always improves when combined with traditional \gls{ssl} methods. \textbf{(v)} Finally, we provide theoretical and empirical insights why such a feature distribution misalignment occurs and show that our method successfully reduces it.

The paper is organized as follows: Sec.~\ref{sec:relatedwork} 
discusses the related work, we define the problem of \gls{ssl} in Sec.~\ref{sec:problem} and propose our method in Sec.~\ref{sec:method}. Then, we describe the experiments in Sec.~\ref{sec:experiments} and report the results in Sec.~\ref{sec:results}. Finally, we analyse our method in Sec.~\ref{sec:analysis} and conclude this paper with Sec.~\ref{sec:conclusion}.
\section{Related Work}\label{sec:relatedwork}

In this section we review existing \gls{ssl} and \gls{uda} works.

\subsection{Semi-Supervised Learning}

A type of \gls{ssl} methods that is among the most popular and most effective is consistency regularization \cite{laine2017temporal,miyato2018vat,tarvainen2017meanteacher,rasmus2015ladder,sajjadi2016regularization}. It is based on the idea that the class prediction of a sample with different small perturbations should be identical and therefore be in the same cluster that corresponds to a specific class category in a high dimensional feature space~\cite{oliver2018realistic}. This idea is implemented using a student and a teacher model, where both obtain the same input samples but with different perturbations but their prediction should agree. Thus, consistency regularization requires to update the student model such that it produces the same prediction as the teacher model. Among the first methods following this idea is the $\Pi$-model~\cite{laine2017temporal} that uses the same model for the student and the teacher but applies different random perturbations on the input image fed to each model. Inconsistency between two models is penalized by computing the mean squared error between both predictions.  Mean-Teacher~\cite{tarvainen2017meanteacher} uses an improved model as teacher, that uses exponential moving average to update the model parameters. Instead of using random perturbations for data augmentation, VAT~\cite{miyato2018vat} generates small adversarial perturbations that deteriorate the class prediction for a given sample the most.

Other \gls{ssl} methods use \textit{predicted} labels of unlabeled samples for training~\cite{cicek2018saas,lee2013pseudo}. As soon as the confidence of the model for a given sample is high enough, this sample and its pseudo label are included into training~\cite{lee2013pseudo}. SAAS~\cite{cicek2018saas} uses the speed of convergence when training with predicted labels, to assess the quality of the model and to improve it accordingly.  

GAN based \gls{ssl} methods~\cite{chongxuan2017triple,dai2017badgan,salimans2016improved,springenberg2015unsupervised} train a generator to produce synthetic unlabeled samples and a discriminator that not only differentiates between real and fake samples, but is also trained to predict the class labels. In contrast, TripleGAN~\cite{chongxuan2017triple} separates the classifier from the discriminator and generates synthetic image/label pairs or predicts labels for unlabeled samples. The task of the discriminator is to distinguish between real and fake image/label pairs. Thus, GAN based \gls{ssl} methods use advanced data augmentation techniques compared to consistency regularization.

Another line of research focuses on methods that are complementary to traditional \gls{ssl} techniques and achieve competitive results when combined with other approaches~\cite{luo2018smooth}, \eg Entropy minimization~\cite{grandvalet2005semi} that is based on the observation that the prediction for each sample is unique. Hence, it adds a penalty if the model produces non-unique predictions.

\subsection{Unsupervised Domain Adaptation}

Popular methods tackling \gls{uda} are mainly based on minimizing the \gls{mmd}~\cite{long2015learning,tzeng2014confusion} between the source and the target distributions or using an adversarial training approach in order to align both distributions. However, since the development of \glspl{gan}~\cite{goodfellow2014generative} more and more adversarial based \gls{uda} methods emerged and continuously improved the SOTA. Ganin~\etal~\cite{ganin15domain} proposed to align images from both domains in a shared feature space. The main idea is training a discriminator besides a feature extractor and a classifier. The discriminator is trained to identify labeled samples using only the extracted features. Thus, confusing the discriminator corresponds to aligned feature distributions. Ganin~\etal~\cite{ganin15domain} proposed a gradient reversal layer in order to train the feature extractor, classifier and discriminator jointly. Based on this work plenty of other methods have emerged in order to better align source and target domains in feature space~\cite{tzeng2017adversarial,pinheiro2018similarity,pan2019prototypical,chen2019progressive}. Tzeng~\etal~\cite{tzeng2017adversarial} proposed an optimization strategy for adversarial based \gls{uda} similar to training \glspl{gan} without requiring a specialized gradient reversal layer.

\section{Problem Definition and Motivation}\label{sec:problem}

In this section, we first formally describe the problem of \gls{ssl} and present the motivation of this paper.
In \gls{ssl} we are given a set of labeled $\mathcal{X}_l = \{ \img^i_l \}_{i=1}^N$ with labels $\mathcal{Y}_l = \{\lab^i_l\}_{i=1}^N$, where $y_l^i = \{1,2,\dots,K\}$ and a set of unlabeled samples $\mathcal{X}_{ul} = \{ \img^i_{ul} \}_{i=1}^M$. Both sets contain \textit{i.i.d.} samples from the set $\mathcal{X} = \{ \img^i \}_{i=1}^{N+M}$. Thus, the label and sample distributions of both types are aligned by definition. However, the labeled set is typically much smaller than the unlabeled set ($N < M$) such that training a deep learning model solely on the labeled samples leads to overfitting. Thus, \gls{ssl} methods~\cite{chapelle2009ssl} make use of the large unlabeled dataset in order to reduce overfitting and improve the accuracy of the classifier $h$ parameterized by $\theta_h$. But first, we introduce the supervised loss that uses only the labeled samples to train the classifier and reads as follows

\begin{equation}
    \min_\theta \loss_\textrm{Sup}(\DX_l, \DY_l, \theta_h) = \min_\theta \frac{1}{N}\sum_{i=1}^{N}\ell(h(x_l^i;\theta_h), y_l^i),
\end{equation}

where $\ell$ is the loss function, \eg cross-entropy loss. Furthermore, we follow the notation of \gls{uda} and decompose the classifier $h$ into $g$ and $f$ such that $h = g \circ f$ where $f$ maps samples into feature space $g$ outputs the predictions given extracted features and $\theta_h = \{\theta_f, \theta_g\}$ parameterize both modules.

\begin{figure}[t]
\centering 
\subfloat[SVHN]{\includegraphics[width=0.5\columnwidth, keepaspectratio]{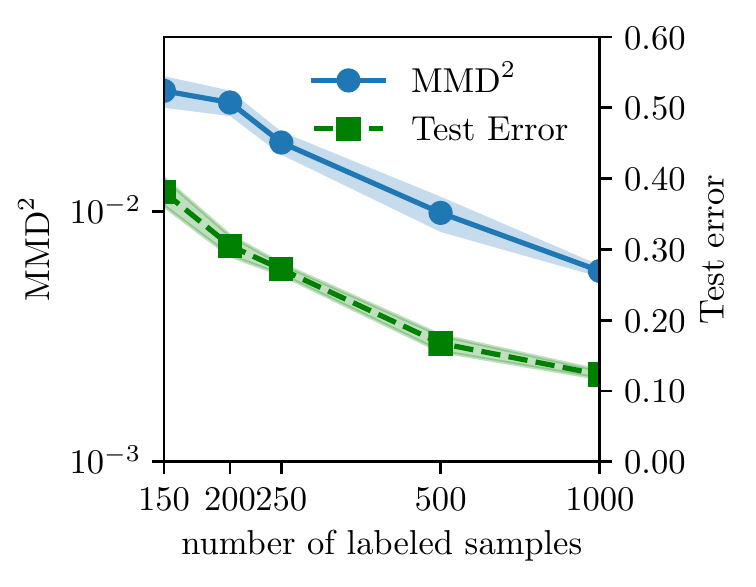}}
\subfloat[CIFAR-10]{\includegraphics[width=0.5\columnwidth, keepaspectratio]{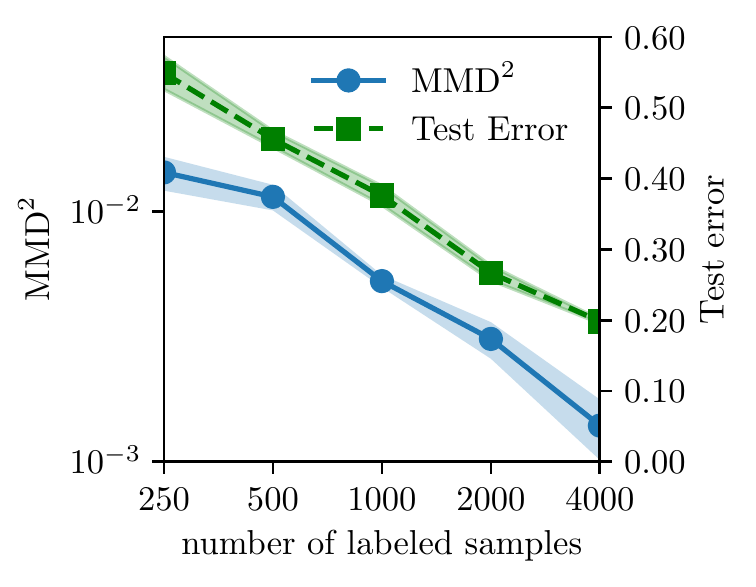}}
\caption{Test error and feature distribution alignment between labeled and unlabeled samples ($\mathrm{MMD}^2$). The dots show the mean and the shaded areas the standard deviation for five runs with different random seeds.}\label{fig:mmd-plots}

\end{figure}

Now, we want to measure the quality of generalization of the trained model and the feature distribution alignment between labeled and unlabeled samples. Hence, we minimize the supervised loss and report the accuracy of the model on a testing set in order to assess the generalization of the predictor. Similarly, we use the \gls{mmd} between the features of all labeled and the features of all unlabeled samples to compute the degree of alignment between both distributions. To avoid biased estimates,  we use the unbiased empirical estimate of the square \gls{mmd}
introduced by \cite{gretton2012kernel}. Fig.~\ref{fig:mmd-plots} shows the test error and the squared \gls{mmd} on two data sets for different amounts of labeled samples. We observe: \textbf{(i)} the test error decreases when increasing the amount of labeled samples \textbf{(ii)} the distributions are tighter aligned when increasing the number of labeled samples (\gls{mmd} decreases) \textbf{(iii)} the test error and \gls{mmd} are strongly correlated. We conclude that small test errors correspond to aligned feature distributions. Thus, the idea of this paper is to actively align the distributions in order to achieve a tighter alignment that corresponds to a smaller test error but without requiring more labeled samples. Thus, we propose in the next section how to overcome feature distribution misalignment for \gls{ssl} by using two different strategies namely adversarial alignment commonly used in \gls{uda} and consistency regularization. 

\begin{figure*}[t]
    \centering
    \includegraphics[width=0.9\textwidth,keepaspectratio]{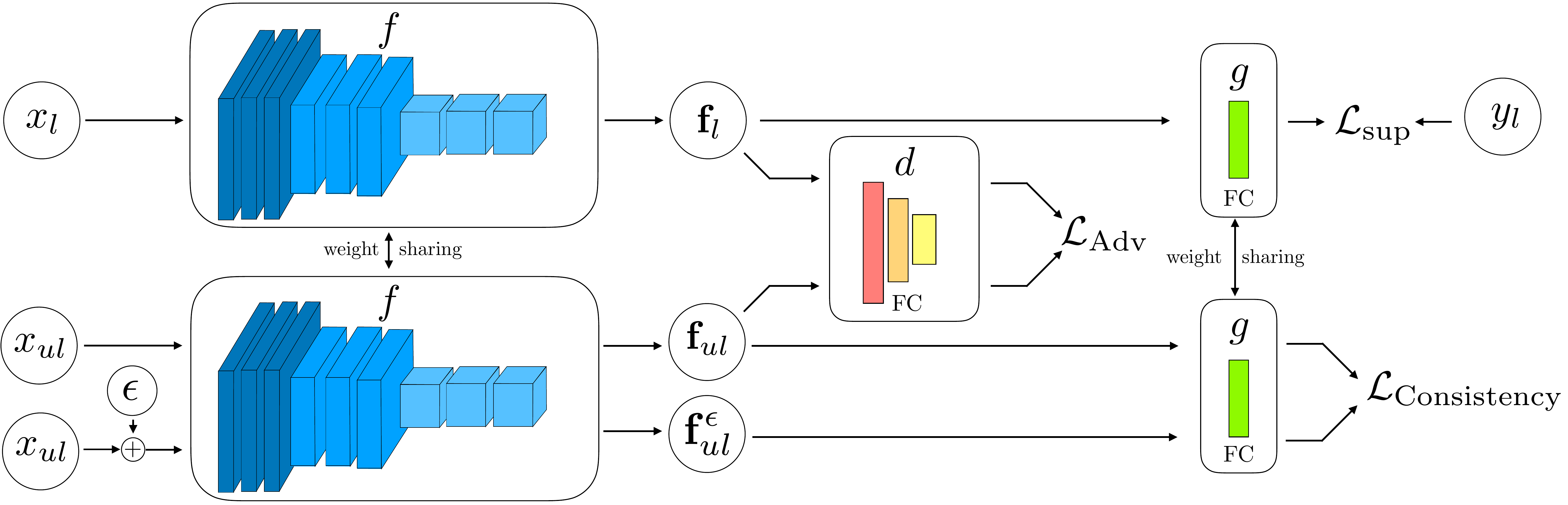}
    \caption{Block diagram of our proposed adversarial feature  distribution alignment method AFDA.}
    \label{fig:block-diagram}
\end{figure*}

\section{Method}\label{sec:method}

In this section we will propose AFDA that overcomes feature distribution misalignment using two different strategies namely adversarial alignment commonly used in \gls{uda} and consistency regularization used in \gls{ssl}.

\subsection{Adversarial Distribution Alignment}

Following the recent trends in \gls{uda} and in training \gls{gan} we propose an adversarial distribution alignment method that aims at minimizing the $\mathcal{H}$-divergence~\cite{ben2010theory}. However, to compute the $\mathcal{H}$-divergence we require an additional model namely a binary discriminator. We train the discriminator $d$ in order to correctly identify whether an extracted feature corresponds to a labeled or an unlabeled sample. As a consequence, the feature distributions are aligned as soon as the discriminator no longer reliably classifies the feature representations. We use the binary cross entropy loss to measure the error of the discriminator. The proposed adversarial loss reads as follows

\begin{multline}\label{eq:ladv}
    \max_{\theta_d}\loss_{\textrm{Adv}}(\DX_l, \DX_{ul}; \theta_d, \theta_f) \coloneqq\\ \max_{\theta_d}\Biggl\{\frac{1}{N}\sum_{i=1 }^{N}\log d(f(x_l^i;\theta_f); \theta_d) \\ +\frac{1}{M}\sum_{i=1}^{M}\log(1 - d(f(x_{ul}^i;\theta_f);\theta_d))\Biggl\},
\end{multline}

where $d$ is the discriminator producing the probability how likely a feature representation corresponds to a labeled or unlabeled sample. The parameters $\theta_d$ and $\theta_f$ parameterize the discriminator $d$ and the feature extractor $f$.

\subsection{Consistency Regularization}

In addition to the proposed adversarial distribution alignment losses we add a consistency regularization loss. The motivation are two-fold. First, the proposed losses only allow to train the feature extractor $f$ and update the parameters $\theta_f$ but not the classifier $g$ with $\theta_g$, see Eq.~\eqref{eq:ladv}. In contrast, consistency regularization requires that the classifier predictions for the same samples with different perturbations are equivalent. Hence, the classifier $g$ is trained with unlabeled samples such that they allow to improve the weights $\theta_g$. Furthermore, adversarial distribution alignment solely aligns the marginal distributions instead of the conditional distributions. The reason is that in \gls{ssl} the amount of labeled samples is much smaller than in \gls{uda} such that only a few samples per category are labeled and the conditional distributions are less representative. Therefore, we propose to use consistency regularization to achieve representative conditional distributions such that aligning the marginal distributions works more reliably. Furthermore, consistency regularization enforces that the same samples with different perturbations achieve a similar representation in the feature space and as a consequence a similar prediction. Thus, it is less likely that distribution alignment assigns samples to a wrong cluster that corresponds to another category in the high dimensional feature space than the sample.

The consistency loss is then defined as follows

\begin{multline}
    \loss_{\textrm{Consistency}}(\DX_{ul};\theta_h) \coloneqq\\\frac{1}{M}\sum_{i=1}^{M} \mathrm{div}(h(x_{ul}^i;\theta_h), h(x_{ul}^i +  \epsilon;\theta_h)),
\end{multline}
where $\epsilon$ is the specific perturbation and $\textrm{div}$ is a distance function, \eg Kullback-Leibler divergence or Euclidean norm.

\subsection{Optimization Algorithm}

Combining all three loss functions results in the following optimization problem and reads as follows

\begin{equation}\label{eq:optimizationproblem}
    \min_{\theta_f,\theta_g}\max_{\theta_d} \loss_\textrm{Sup}(\theta_f, \theta_g) + \mu \loss_{\textrm{Adv}}(\theta_f, \theta_d) + \eta\loss_{\textrm{Cons}}(\theta_f, \theta_g),
\end{equation}

where $\eta$ and $\mu$ are scalars.
Similar optimization problems to Eq.~\eqref{eq:optimizationproblem} occur in the literature of \gls{uda}~\cite{ganin15domain,tzeng2017adversarial} or GANs~\cite{goodfellow2014generative}. Instead of using a gradient reversal layer~\cite{ganin15domain} we solve it in an alternating fashion~\cite{tzeng2017adversarial}, see Alg.~\ref{alg:optimization}. First, we fix the parameters $\theta_d$ while updating the parameters $\theta_f$ and $\theta_g$ using gradient decent. Second, we fix the parameters $\theta_f$ and $\theta_g$ and update the parameters $\theta_d$ using gradient ascent (note the plus sign when updating $\theta_d$ in contrast to the minus signs when updating $\theta_f$ and $\theta_g$). For $\mu$ we follow the schedule of \cite{tarvainen2017meanteacher,laine2017temporal,miyato2018vat} depending on the distance function $\mathrm{dev}$. The schedule to compute $\mu$ reads as follows

\begin{equation}
    \mu(t, T, \lambda) \coloneqq \frac{1 - \exp\left(\lambda t/T\right)}{1 + \exp\left(\lambda t/T\right)},
\end{equation}

where $t$ is the current epoch, $T$ the number of training epochs and $\lambda$ is a hyper parameter.

\RestyleAlgo{boxruled}
\begin{algorithm}[t]
    \KwData{training data sets $\DX_l, \DY_l, \DX_{ul}$}
    \KwResult{the trained network using SSL}
    random initialization: $\theta_f,\theta_g,\theta_d$\;
    initialization: $\lambda, T, \gamma$\\
    \For{$t=0$ \KwTo T}
    {
        start the new epoch $t$\\
        sample batch from $(\DX_l, \DY_l, \DX_{ul})$\\
        $\theta_f^{t+1} = \theta_f^t - \gamma\nabla_{\theta_f}\Big(\loss_\textrm{sup}(\theta_f^t, \theta_g^t)$\\
        \qquad$\mu^t\loss_{\textrm{Adv}}(\theta_f^t, \theta_d^t) + \eta^t\loss_{\textrm{Con}}(\theta_f^t, \theta_g^t)\Big) + $\\
        $\theta_g^{t+1} = \theta_g^t - \gamma\nabla_{\theta_g}\Big(\loss_\textrm{sup}(\theta_f^{t+1}, \theta_g^t)$ + \\
        \qquad$\eta^t\loss_{\textrm{Con}}(\theta_f^{t+1}, \theta_d^t)\Big)$\\
        $\theta_d^{t+1} = \theta_d^t + \gamma\nabla_{\theta_d}\loss_{\textrm{Adv}}(\theta_f^{t+1}, \theta_d^t)$\\
        $\mu^{t+1} = \mu(t+1,T,\lambda)$\\
        $\eta^{t+1} = \eta(t+1)$\\
        update optimizer related parameters ($\gamma, \dots$)
	}
    \caption{Optimization algorithm to solve Eq.~\eqref{eq:optimizationproblem}.}
    \label{alg:optimization}
\end{algorithm}

\subsection{Comparison with SSL based UDA Methods}

To the best of our knowledge, there is only one work that combines \gls{ssl} and domain adaptation but aims at a very different goal.
French~\etal~\cite{french2018selfensembling} use consistency regularization to improve the performance of domain adaptation, \ie they use a large labeled data set sampled from the source domain and test on the target domain, and additionally require consistent predictions for the data points sampled from the target domain. 
Hence, we use distribution alignment for \gls{ssl} and French~\etal~\cite{french2018selfensembling} use \gls{ssl} for unsupervised domain adaptation.
\section{Experiments}\label{sec:experiments}

To demonstrate the effectiveness of our proposed method, we evaluate it on the two most frequently used \gls{ssl} data sets: SVHN~\cite{netzer2011svhn} and CIFAR-10~\cite{krizhevsky2009cifar}.
We implement the widely used CNN architectures ConvLarge~\cite{miyato2018vat} and ResNet with shake-shake regularization~\cite{gastaldi2017shake} that achieve the current SOTA in \gls{ssl}~\cite{yu2019tangent, robert2018hybridnet}.

First, we describe the settings for our proposed method on both data sets. Then, we describe the configurations of multiple additional experiments that we run in order to understand our method better. We report the benefit of using different components of our method or replace our consistency loss by popular losses from the literature in order to show that adding adversarial feature alignment always helps.

\paragraph{Data Sets and Configurations}
We use two different operating points that use between 250 and 4000 labeled samples depending on the data set. We randomly construct five labeled data sets for each operating point and use all other samples for the unlabeled data sets. We rerun each experiment using five random seeds and report mean and standard deviation on the test set for all experiments.
On SVHN we use the Kullback-Leibler divergence and adversarial noise for consistency regularization and add  entropy loss for unlabeled samples~\cite{miyato2018vat}. We train for 300 epochs with randomly sampled labeled batches and unlabeled batches.
On CIFAR-10 we use the euclidean norm and standard data augmentation (translation and vertical flipping) for consistency regularization. We train for 300 epochs with randomly sampled labeled batches and unlabeled batches.

In order to compute the adversarial loss, we extract the features directly after the global average pooling operation or in front of the fully connected layer that produces the logits. Then, we feed these features into the discriminator that consists of a simple three layer perceptron. 
We compute the distribution alignment loss by using only the current batches of labeled and unlabeled samples in each training step.

More details about the evaluation environment, data sets, model architectures and all other parameters are described in the supplementary or in the corresponding papers.

\paragraph{Configuration for the Ablation Study}

For the ablation studies we train our method with or w/o the distributions alignment loss and with or w/o the consistency loss. 

Furthermore, we replicate three popular \gls{ssl} methods: $\Pi$-Model, Mean-Teacher and VAT. This allows combining each method with ours and highlighting the substantial benefit of adding distribution alignment to state-of-the-art \gls{ssl} methods. Furthermore, replicating allows to compare on more operating points, especially on the rarely tested with only a few labels, and a unified evaluation environment ensures that the comparison is fair for all methods~\cite{oliver2018realistic}. We use ConvLarge for both data when running experiments for the ablation study.

\begin{table}[t]
  \renewcommand{\arraystretch}{1.2}
  \newcommand{\head}[1]{\textnormal{\textbf{#1}}}
  \newcommand{\normal}[1]{\multicolumn{1}{c}{#1}}
  \newcommand{\na}{\emph{n.a.}}

  \colorlet{tableheadcolor}{gray!25} 
  \newcommand{\headcol}{\rowcolor{tableheadcolor}} %
  \colorlet{tablerowcolor}{gray!10} 
  \newcommand{\rowcol}{\rowcolor{tablerowcolor}} %
  \setlength{\aboverulesep}{0pt}
  \setlength{\belowrulesep}{0pt}

  \newcommand*{\rulefiller}{
    \arrayrulecolor{tableheadcolor}
    \specialrule{\heavyrulewidth}{0pt}{-\heavyrulewidth}
    \arrayrulecolor{black}}
  
    \centering
  \caption{Test errors (\%) on SVHN using ConvLarge.}
  \label{tab:svhn-comparison-literature}
  \begin{tabular}{lrr}
    \toprule
    \headcol                                            & 250 labels               & 1000 labels      \\
    \toprule
    \head{Supervised-only}                              & $26.04 \pm 1.43$         & $10.74 \pm 0.28$ \\
    \midrule
    \head{$\mathbf{\Pi}$ Model}~\cite{laine2017temporal}&  ---                     & $4.82 \pm 0.17$\\
    \head{TempEns}~\cite{laine2017temporal}             &  ---                     & $4.42 \pm 0.16$\\
    \head{MT}~\cite{tarvainen2017meanteacher}                 & $4.35 \pm 0.50$          & $3.95 \pm 0.19$\\
    \head{ICT}~\cite{verma2019interpolation}            & $4.78 \pm 0.68$          & $3.89 \pm 0.04$\\
    \head{VAT+Ent}~\cite{miyato2018vat}                 & ---                      & $3.86 \pm 0.11$\\
    \head{MT+SNTG}~\cite{luo2018smooth}                 & $4.29 \pm 0.23$          & $3.86 \pm 0.27$\\
    \head{$\mathbf{\Pi}$+SNTG}~\cite{luo2018smooth}     & $5.07 \pm 0.25$          & $3.82 \pm 0.25$\\
    \head{SAAS}~\cite{cicek2018saas}                    & ---                      & $3.82 \pm 0.09$\\
    \head{TNAR+VAE}~\cite{yu2019tangent}                & ---                      & $3.74 \pm 0.04$\\
    \midrule
    \head{AFDA~(Ours)}                  & $\mathbf{3.88 \pm 0.13}$     & $\mathbf{3.39 \pm 0.12}$\\
    \bottomrule
  \end{tabular}
\end{table}
\begin{table}[t]
  \renewcommand{\arraystretch}{1.2}
  \newcommand{\head}[1]{\textnormal{\textbf{#1}}}
  \newcommand{\normal}[1]{\multicolumn{1}{c}{#1}}
  \newcommand{\na}{\emph{n.a.}}

  \colorlet{tableheadcolor}{gray!25} 
  \newcommand{\headcol}{\rowcolor{tableheadcolor}} %
  \colorlet{tablerowcolor}{gray!10} 
  \newcommand{\rowcol}{\rowcolor{tablerowcolor}} %
  \setlength{\aboverulesep}{0pt}
  \setlength{\belowrulesep}{0pt}

  \newcommand*{\rulefiller}{
    \arrayrulecolor{tableheadcolor}
    \specialrule{\heavyrulewidth}{0pt}{-\heavyrulewidth}
    \arrayrulecolor{black}}
  
    \centering
  \caption{Test errors (\%) on CIFAR-10 using shake-shake ResNet or ConvLarge where testing on ResNet is omitted.}
  \label{tab:cifar-comparison-literature}
  \begin{tabular}{lrr}
    \toprule
    \headcol                                            & 1000 labels              & 4000 labels      \\
    \toprule
    \head{VAT+Ent}~\cite{miyato2018vat}                 & ---                      & $10.55 \pm 0.05$\\
    \head{TEns+SNTG}~\cite{luo2018smooth}               & $18.41 \pm 0.52$          & $10.93 \pm 0.14$\\
    \head{TNAR+VAE}~\cite{yu2019tangent}                & ---                      & $8.85 \pm 0.03$\\
	\head{ICT}~\cite{verma2019interpolation}            & $15.48 \pm 0.78$          & $7.29 \pm 0.02$\\
    \midrule
    \midrule
    \head{Supervised-only}                              & $45.20 \pm \textrm{n.a.}$ & $15.45 \pm \textrm{n.a.}$ \\
    \midrule
    \head{MT}~\cite{tarvainen2017meanteacher}                 & $10.08 \pm 0.41$        & $6.28 \pm 0.15$\\
    \head{HybridNet}~\cite{robert2018hybridnet}              & $ \mathbf{8.81 \pm \textrm{n.a.}}$        & $6.09 \pm \textrm{n.a.}$\\
    \midrule
    \head{AFDA~(Ours)}                       & $9.40 \pm 0.32$          & $\mathbf{6.05 \pm 0.13}$\\
    \bottomrule
  \end{tabular}
\end{table}

\section{Results}\label{sec:results}

First, we compare to recent methods in the literature on both data sets. Then, we show the results of our ablation study where we assess the performance with and w/o particular losses used for training the model.

The results in Tab.~\ref{tab:svhn-comparison-literature} demonstrate that our method achieves the SOTA on SVHN. Furthermore, we want to highlight that the test error of $3.39\%$ is only marginally higher than $2.89\%$, the error of the fully supervised method that uses all 73k labeled samples for training. Similarly, Tab.~\ref{tab:cifar-comparison-literature} shows our method achieves state-of-the-art results on CIFAR-10. Note, that we avoid to exhaustively list every \gls{ssl} method present in the literature. We rather concentrate on the most similar methods that use the same architectures, training efforts and avoid advanced data augmentation.   

Tab.~\ref{tab:improvement-summary} reports the test error when using classical consistency regularization methods such as VAT, Mean-Teacher or $\Pi$-model with and w/o adversarial alignment. We observe that adding adversarial alignment always reduces the test error, except for one operating point where both are equivalent. The reduction is even more significant when using only a few labeled samples. For 200 labeled samples, adding adversarial alignment to VAT achieves a test error of $4.91\%$ which is smaller than any error reported in Tab.~\ref{tab:improvement-summary} that is not using adversarial alignment.

The results in Tab.~\ref{tab:improvement-summary} demonstrate that only using adversarial alignment achieves already a smaller test error than the supervised-only method on each operating point. Furthermore, adversarial alignment clearly outperforms
state-of-the-art \gls{ssl} methods when using only a small amount of labelled samples. Adversarial alignment without consistency achieves a test error of $9.09\%$ using 150 labeled samples, that corresponds to a reduction of $27.72$ ($4\times$ relative reduction) compared to the baseline, that uses only the labeled samples for training. In contrast, the runner up (VAT) achieves only a test error of $18.3\%$ ($2\times$ relative reduction) using the same amount of labeled samples.

\begin{figure*}[t]
\centering 
\subfloat[
Supervised-only\newline\hspace*{1.5em}
$\mathrm{MMD}^2$: $27.3 \pm 3.2$\newline\hspace*{1.5em}
Test Error: $29.8 \pm 1.4$%
]{\includegraphics[width=0.5\columnwidth, keepaspectratio]{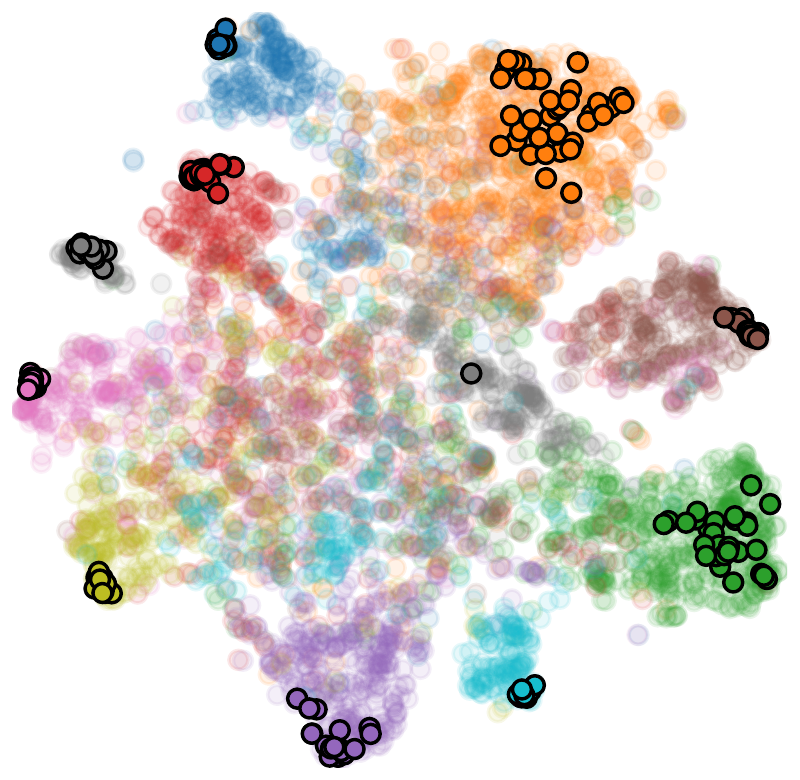}\label{fig:tsne-ablation-sup}}
\subfloat[
AFDA w/o Alignment\newline\hspace*{1.5em}
$\mathrm{MMD}^2$: $13.7 \pm 1.9$\newline\hspace*{1.5em}
Test Error: $13.4 \pm 1.6$ 
]{\includegraphics[width=0.5\columnwidth,keepaspectratio]{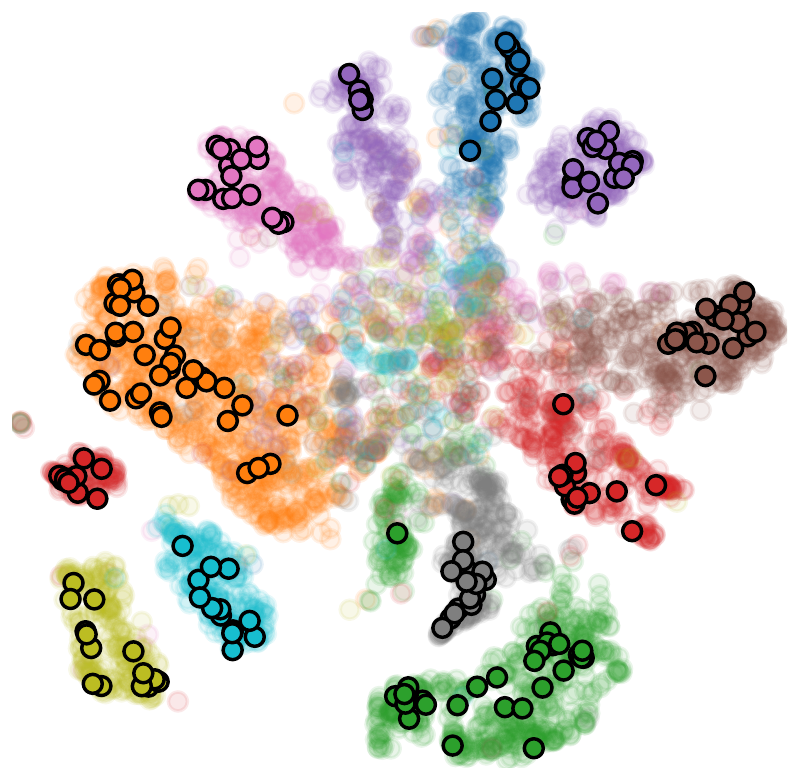}\label{fig:tsne-ablation-con}}
\subfloat[
AFDA w/o Consistency\newline\hspace*{1.5em}
$\mathrm{MMD}^2$: $1.1 \pm 0.5$\newline\hspace*{1.5em}
Test Error: $8.7 \pm 0.5$ 
]{\includegraphics[width=0.5\columnwidth, keepaspectratio]{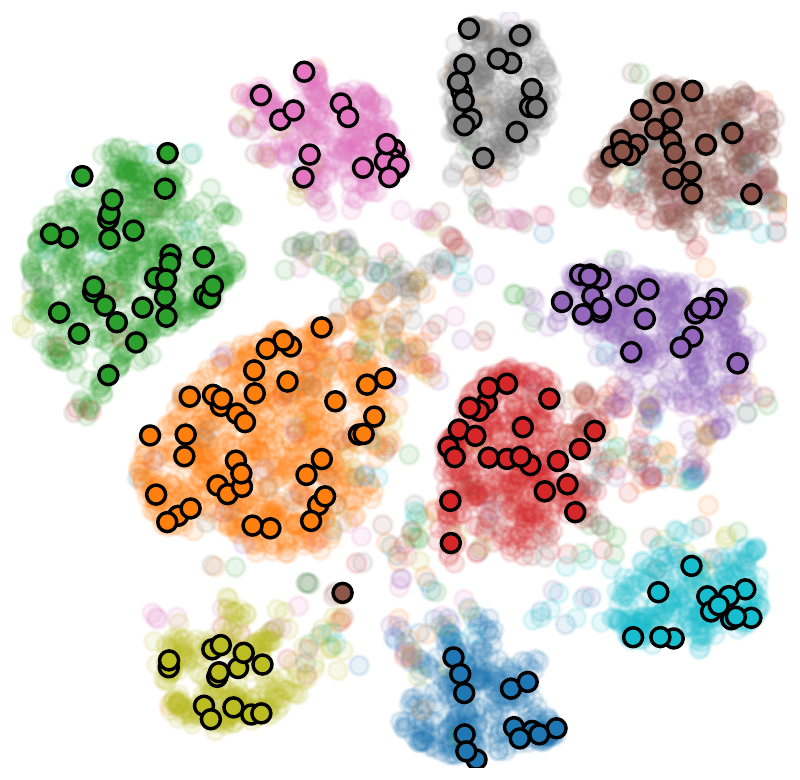}\label{fig:tsne-ablation-dom}}
\subfloat[
AFDA with Cons. and Align.\newline\hspace*{1.5em}
$\mathrm{MMD}^2$: $2.3 \pm 0.4$\newline\hspace*{1.5em}
Test Error: $4.9 \pm 0.1$ 
]{\includegraphics[width=0.5\columnwidth, keepaspectratio]{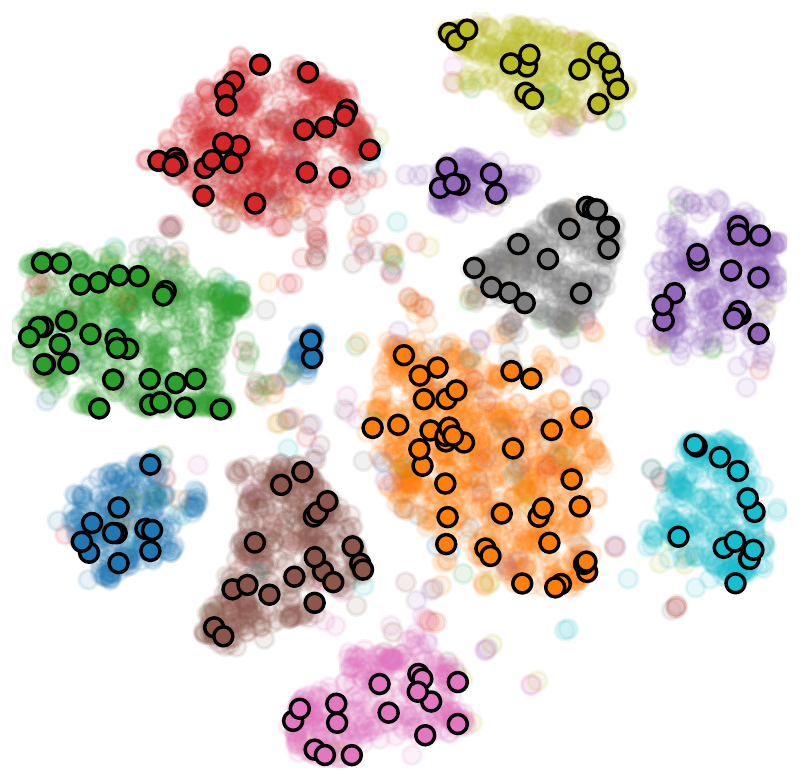}\label{fig:tsne-ablation-dom-con}}

\caption{T-SNE of the feature space when using different losses to train AFDA. The disks with a black border correspond to labeled and all other to unlabeled samples, the colors identify the class. We use 200 labeled SVHN samples. The plots show 200 labeled and 5000 unlabeled samples. The reported $\mathrm{MMD}^2$ is scaled by $\times 10^{3}$ and the test error is reported in \%. Best viewed in color.}\label{fig:tsne-ablation}

\end{figure*}

In Fig.~\ref{fig:tsne-ablation} we report our method when excluding the adversarial alignment loss and/or the consistency loss. Fig.~\ref{fig:tsne-ablation} shows the t-SNE embedding of the feature space, $\mathrm{MMD}^2$ between the feature distributions corresponding to labeled and unlabeled samples and the test error of the trained model on the test set. We observe that only using the supervised loss in Fig.~\ref{fig:tsne-ablation-sup} leads to overfitting such that the model groups labeled samples close together (circles with black boundary) but generalizes poorly to unlabeled samples (circles without black boundary). The two feature distributions are not aligned as labeled samples are missing in the center of the embedding. The high $\mathrm{MMD}^2$ is another indicator for poor alignment.
When adding th consistency loss (see Fig.~\ref{fig:tsne-ablation-con}) we visually and empirically ($\mathrm{MMD}^2$) observe a tighter alignment between both distributions. Thus, consistency regularization helps but there still exists a region in the embedding where only unlabeled samples occur. 
Fig.~\ref{fig:tsne-ablation-dom} shows that ours without consistency loss results in clearly defined clusters for each class and tightly aligned distributions in feature space. Identifying the origin of a sample solely by the feature representation is no longer possible, but alignment is still not optimal.
Using all losses results in a feature embedding where even more samples are clearly assigned to a cluster and achieves the smallest test error, see Fig,~\ref{fig:tsne-ablation-dom-con}.

\begin{table*}[t]
  \renewcommand{\arraystretch}{1.2}
  \newcommand{\head}[1]{\textnormal{\textbf{#1}}}
  \newcommand{\normal}[1]{\multicolumn{1}{c}{#1}}
  \newcommand{\na}{\emph{n.a.}}

  \colorlet{tableheadcolor}{gray!25} 
  \newcommand{\headcol}{\rowcolor{tableheadcolor}} %
  \colorlet{tablerowcolor}{gray!10} 
  \newcommand{\rowcol}{\rowcolor{tablerowcolor}} %
  \setlength{\aboverulesep}{0pt}
  \setlength{\belowrulesep}{0pt}

  \newcommand*{\rulefiller}{
    \arrayrulecolor{tableheadcolor}
    \specialrule{\heavyrulewidth}{0pt}{-\heavyrulewidth}
    \arrayrulecolor{black}}
  
    \centering
  \caption{Test errors (\%) for different \gls{ssl} methods with and without adversarial distribution alignment.}
  \label{tab:improvement-summary}

\resizebox{1.\textwidth}{!}{%
  \begin{tabular}{lc*{8}{r}}
    \toprule
    \headcol                                                         &          & \multicolumn{3}{c}{SVHN ($2.89 \pm 0.05$ with 73k labels)} & \multicolumn{3}{c}{CIFAR-10 ($6.87 \pm 0.14$ with 50k labels)}\\
    \cmidrule(lr){3-5}\cmidrule(lr){6-8}
    \headcol                                                         &   $\loss_\textrm{Adv}$       & 150 labels          & 200 labels        & 1000 labels      & 250 labels         & 500 labels       & 4000 labels   \\
    \toprule
     \multirow{2}{*}{\head{No Consistency}}       & ---        & $36.81 \pm 3.08$ & $29.82 \pm 1.37$  & $10.74 \pm 0.28$  & $54.69 \pm 1.32$ & $45.69 \pm 1.37$ & $19.74 \pm 0.28$  \\
                                                  & \checkmark & $9.09  \pm 1.08$ & $8.72  \pm 0.53$  & $7.54  \pm 0.26$  & $43.79 \pm 1.67$ & $35.36 \pm 1.09$ & $15.55 \pm 0.37$  \\
    \midrule
    Error Reduction                                     &            & 27.72            & 21.10             & 3.20              & 10.90            & 10.34            & 4.19              \\
    \midrule
    \midrule
    \multirow{2}{*}{\head{$\mathbf{\Pi}$ Model}} & ---        & $31.23 \pm 1.93$ & $17.80 \pm 1.84$   & $6.12 \pm 0.24$   & $53.68 \pm 1.50$ & $44.41 \pm 0.79$ & $15.98 \pm 0.16$  \\
                                                 & \checkmark & $8.62 \pm 0.36$  & $6.96 \pm 0.71$    & $5.16 \pm 0.28$   & $43.42 \pm 2.03$ & $34.00 \pm 1.11$ & $13.64 \pm 0.29$  \\
    \midrule                                             
    Error Reduction                                    &            & 22.61            & 8.84               & 0.96              & 10.26            & 10.41            & 2.34              \\
    \midrule
    \midrule
    \multirow{2}{*}{\head{Mean Teacher}}         & ---        & $29.69 \pm 7.92$ & $15.62 \pm 0.97$   & $5.03 \pm 0.19$   & $53.09 \pm 1.60$ & $44.47 \pm 1.13$ & $15.35 \pm 0.18$  \\
                                                 & \checkmark & $\mathbf{7.48 \pm 1.25}$  & $6.93 \pm 0.70$    & $5.14 \pm 0.20$   & $41.75 \pm 1.52$ & $33.27 \pm 0.82$ & $13.69 \pm 0.23$  \\
    \midrule
    Error Reduction                                    &            & 12.21            & 8.69               & -0.09             & 11.34            & 11.20            & 1.67              \\ 
    \midrule
    \midrule
    \multirow{2}{*}{\head{VAT}}                  & ---        & $18.30 \pm 4.02$ & $13.41 \pm 1.55$   & $5.91 \pm 0.37$   & $39.90 \pm 3.73$ & $27.95 \pm 3.56$ & $12.59 \pm 0.21$  \\
                                                 & \checkmark & $8.93 \pm 3.24$  & $\mathbf{4.91 \pm 0.12}$    & $\mathbf{4.93 \pm 0.07}$   & $\mathbf{29.44 \pm 2.00}$ & $\mathbf{20.47 \pm 0.66}$ & $\mathbf{11.54 \pm 0.16}$  \\
    \midrule
    Error Reduction                                    &            & 9.63             & 8.50               & 0.98              & 10.46            & 7.48             & 1.05              \\
    \bottomrule
  \end{tabular}
  }
\end{table*}

\section{Analysis}\label{sec:analysis}

Whereas we have shown in the previous sections that feature distribution misalignment is a problem in \gls{ssl}, and that our proposed method is able to align them, it is still not entirely clear why such a misalignment is possible. Thus, in this section, we attempt to provide a theoretical insight in order to analyze this phenomenon.

In order to assess the degree of overfitting or the generalization quality, we aim at computing the generalization error that measures the difference between the empirical risk $\hat{R}$ on the training set and the expected risk $R$.
As we do not have access to the true data distribution, it is unfeasible to compute the expected risk. Hence, we approximate it by the empirical risk on a large enough data set. According to Hoeffdings inequality we can bound the difference with respect to the number of samples $n$ and probability at least $1-\delta$~\cite{hoeffding1994probability,bousquet2003introduction}. Thus, we follow Sener~\etal~\cite{sener2018} and use the large amount of unlabeled samples to approximate the expected risk and aim at minimizing the term $|\hat{R}(\DX_{ul}) - \hat{R}(\DX_l)|$ in order to reduce the generalization error 
(note, that the approximation is only valid as long as we are not using the annotations of the unlabeled samples during training).

\begin{theorem}

Let $\W$, $\bias$ be the weight matrix and the bias vector of a fully-connected layer where $\W_k$ denotes the $k^{\text{\small th}}$ row vector and $\bias_k$ the $k^{\text{\small th}}$ scalar, then the difference between the expected risks $|\hat{R}(\DX_{ul}) - \hat{R}(\DX_{l})|$ using the cross entropy loss can approximately be bounded by

\begin{align}
    |\hat{R}(\DX_{ul}) - &\hat{R}(\DX_{l})|
    \lessapprox\sum_{k=1}^{K}\Biggl[ \nonumber \\
    \abs{\W_k}&\abs{\frac{1}{N}\sum_{i=1}^N \delta_k(y_l^i)f(x_{l}^j) - \frac{1}{M}\sum_{i=1}^M \delta_k(y_{ul}^i) f(x_{ul}^i)} \label{eq:feats-true}\\
    + \abs{\bias_k}&\abs{\frac{1}{N}\sum_{i=1}^N \delta_k(y_l^i) - \frac{1}{M}\sum_{i=1}^N \delta_k(y_{ul}^i)} \label{eq:label-true}\\
    +\abs{\W_k}&\abs{\frac{1}{N}\sum_{i=1}^N \delta_k(h(x_l^i))f(x_{l}^j) - \frac{1}{M}\sum_{i=1}^M \delta_k(h(x_{ul}^i)) f(x_{ul}^i)}\label{eq:feats-pred}\\
    + \abs{\bias_k}&\abs{\frac{1}{N}\sum_{i=1}^N \delta_k(h(x_l^i)) - \frac{1}{M}\sum_{i=1}^N \delta_k(h(x_{ul}^i))}\Biggl],\label{eq:label-pred}
\end{align}
where $\delta_p(q)$ is one if $p=q$ and zero otherwise and the operator $|\cdot|$ returns the absolute value element-wise for vectors.
\end{theorem}

\noindent \emph{Proof} See Supplementary.\\

This bound consists of four different terms that measure the feature and label distributions between labeled and unlabeled samples. Thus, reducing the approximate generalization error $|\hat{R}(\DX_{ul}) - \hat{R}(\DX_{l})|$ requires minimizing each term. Term \eqref{eq:label-true} and \eqref{eq:label-pred} measure the similarity of the true and predicted label distributions between the labeled and unlabeled data sets. Term \eqref{eq:feats-true} measures for each category the distance between the mean feature representation of labeled and unlabeled samples. Term \eqref{eq:feats-pred} is similar to \eqref{eq:feats-true} but the predicted instead of the true labels define the mean feature representation per category. 
Note, that prior training on the labeled data set prevents trivial solutions such as assigning each sample the same category or random assignments.

Thus, Theorem 1 shows that a distribution misalignment is indeed possible even though both data sets are sampled from the same data distribution.  Furthermore, we conclude, that misaligned feature and label distributions are a direct consequence of overfitting, \ie the hypothesis $h$ that minimizes the empirical risk $\hat{R}(\DX_l)$ generalizes only poorly to the unlabeled data set.

\begin{figure}[t]
\centering 
\includegraphics[width=0.9\columnwidth,keepaspectratio]{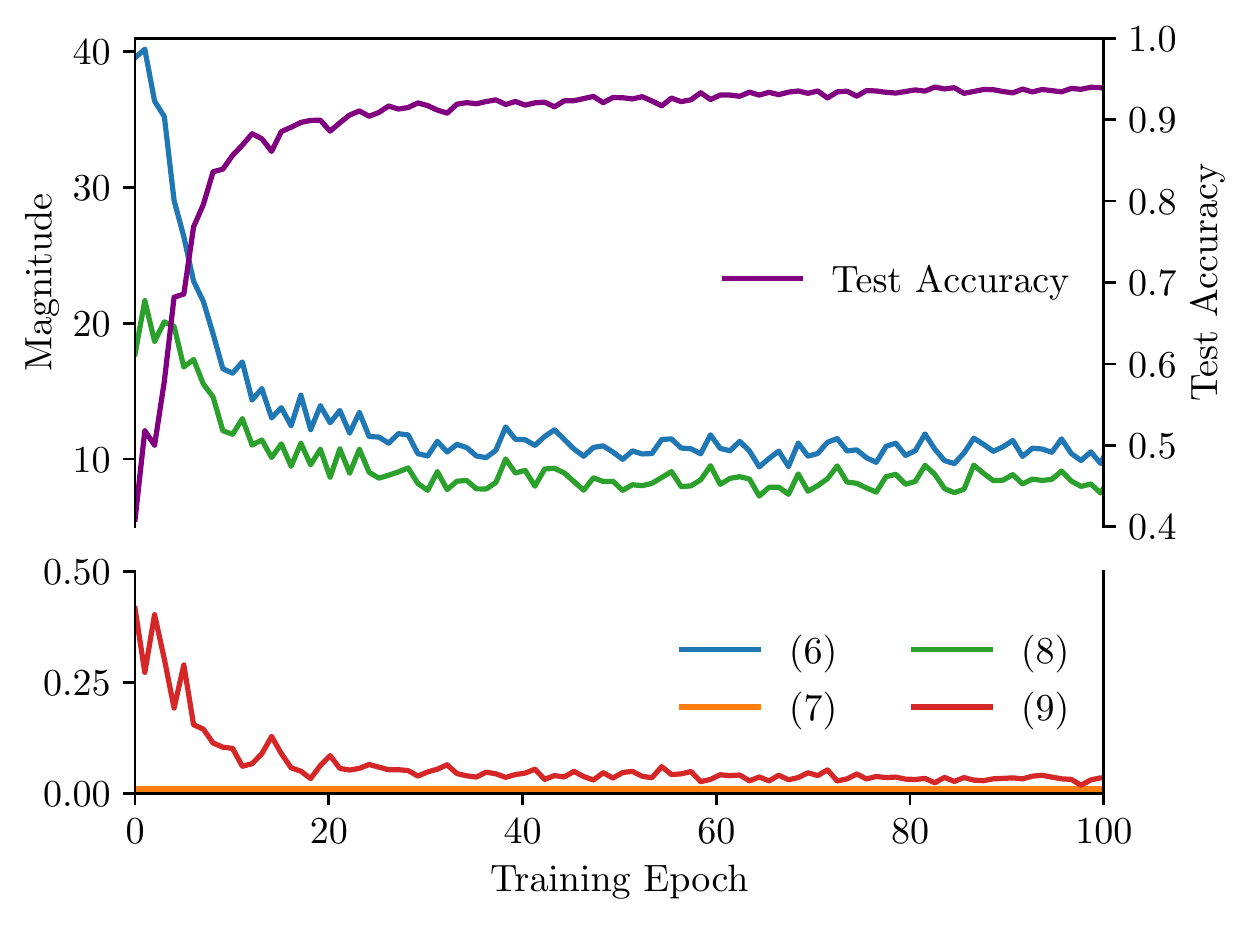}
\caption{Magnitudes of each term that bound the approximation of the generalization error and the test accuracy when training AFDA on SVHN with 250 labeled samples for 100 epochs. Best viewed in color.}\label{fig:bounds-svhn-domain}
\end{figure}

Furthermore, Fig.~\ref{fig:bounds-svhn-domain} shows the values of the terms~\eqref{eq:feats-true} - \eqref{eq:label-pred} when training on SVHN with 250 labeled samples. We observe that our method is able to decrease all four terms. While the terms decrease the test accuracy increases. 

\section{Conclusion}\label{sec:conclusion}

In this paper, we first demonstrated that using only a small amount of labeled samples leads to poor generalization and causes misaligned feature distributions between labeled and unlabeled samples. Thus, we presented a new and very effective \gls{ssl} method that is based on an adversarial feature distribution alignment loss and consistency regularization. The results of the experiments demonstrate that AFDA is particularly suitable when only a small amount of labeled samples is available. Furthermore, we observed that adding adversarial alignment to traditional consistency based \gls{ssl} methods always improves, and leads to state-of-the-art results on SVHN and CIFAR-10. Finally, we provide additional by a theoretical bound for risk minimization and show that AFDA reduces it.

{\small
\bibliographystyle{ieee_fullname}
\bibliography{references.bib}
}
\onecolumn
\appendix
\section{Supplementary}

\subsection{Data Sets, Preprocessing and Augmentation}

We use the most commonly used data sets in \gls{ssl}: SVHN~\cite{netzer2011svhn} and CIFAR-10~\cite{krizhevsky2009cifar}.
SVHN consists of almost 100k 32x32 RGB images of house numbers. Each digit represents a class. The data set used is split between 73k training images and 26k test images.
We rescale the pixels of all images to the range $[-0.5, 0.5]$, whereas augmentation consists of random translations of two pixels in each direction.
CIFAR-10 consists of 60k RGB images with resolution of 32x32 pixels, 10 classes and with 6000 images per class. The data set is split between 50k training images and 10k test images.
The classes represented are: airplanes, cars, birds, cats, deer, dogs, frogs, horses, ships, and trucks. We follow \cite{laine2017temporal,tarvainen2017meanteacher} and use ZCA normalization, whereas augmentation consists of random translations by two pixels and random horizontal flips.  

\subsection{Model Architectures}

The architecture of the classifier $h(\img) = (g \circ f)(\img)$ are equivalent to the classifier introduced by~\cite{miyato2018vat}. The feature extractor $f$ consists of all layers except the last fully connected layer. Therefore, we extract the features directly after global average pooling. Then, the feature extractor produces for each input sample a vector containing $128$ features. 
Furthermore, the shallow classifier $g$ contains only one fully connected layer to match the original classifier proposed by~\cite{miyato2018vat}. Hence, the shallow classifier requires only a small amount of labeled samples for training. 
The discriminator $d$ is a three layer fully-connected neural network using $\mathrm{ReLU}$ activation without batch normalization. The architecture looks as follows: $128 \rightarrow 1024 \rightarrow \mathrm{ReLU} \rightarrow 1024  \rightarrow \mathrm{ReLU} \rightarrow 1 \rightarrow \mathrm{sigmoid}$.

The ResNet architecture with shake-shake regularization is equivalent to the one proposed by~\cite{tarvainen2017meanteacher}. We extract the features just before the last fully connected layer. Thus, the classifier is this fully connected layer and the discriminator is again a three layer perceptron that looks as follows $384 \rightarrow 1024 \rightarrow \mathrm{ReLU} \rightarrow 1024  \rightarrow \mathrm{ReLU} \rightarrow 1 \rightarrow \mathrm{sigmoid}$. We use exactly the same training parameters as proposed by~\cite{tarvainen2017meanteacher}.

In the supplementary, we provide additional experiments using a wide residual network. We use exactly the same architecture as~\cite{oliver2018realistic} namely the WRN-28-2. We use all but the last fully connected layer as the feature extractor. Besides, the last step in the feature extractor is a global average pooling such that the feature vector contains 128 entries. Therefore, we use exactly the same discriminator and the same shallow classifier as described before.

\subsection{Training and Testing Environments}

All networks are trained for 60k iterations for SVHN and 150k iterations for CIFAR-10. We use for all \gls{ssl} methods the same batch size of 100 samples. The samples are selected from the labeled and unlabeled data sets. We use the validation set to tune the initial learning rate, method specific hyper-parameters and the number of labeled training samples in a batch. We observed, that different methods require different fractions of labeled and unlabeled samples in the batches. 
To tune all the parameters, we construct different validation sets sampled from the training set. We use reasonably large validation sets by adapting their size based on the training set. The size of the validation set is as big as the training set but contains at most 1000 labeled samples, i.e. 250 labeled samples for both sets or 4000 in the training and 1000 in the validation set.

Unless stated otherwise, we use the following setting for all experiments: Adam optimizer, linear learning rate decay for the last 20k iterations that decays the learning rate by 0.0025 each 500 iterations, we change $\beta_1$ of the optimizer for the last 20k iterations from 0.9 to 0.5. We follow~\cite{miyato2018vat} for all these choices. Furthermore, we use the validation set to select the amount of labeled samples in the training batches. The searching space is the set $\{2,4,8,16,32\}$. In addition, we tune the initial learning rate from the set $\{0.0001, 0.0005, 0.001, 0.005\}$. Note, that we tune the hyper parameters individually, for all methods and operating points. We observed that we obtain in most cases the same learning rate of 0.0005. In contrast, we observed, that the number of labeled samples per batch depends on the size of the labeled data set. We achieve better performance when using less labeled samples per batch when a small amount of labeled samples is provided, and use more labeled samples per batch when using a large amount of labeled samples for training.

For achieving the new SOTA on SVHN stated in the main paper, we use the same environment as explained above but train for 300 epochs with 32 labeled and 128 unlabeled samples per batch and with a learning rate of 0.0005. 

\subsubsection{Fully-Supervised Model}

We use the aforementioned setup to train the fully-supervised model except that we use only the labeled samples for training and minimize the loss $\loss_\textrm{sup}$. We test different batch sizes instead of 100 because we observed poor performance with such a large batch size when using only a slightly larger training set. Therefore, we use the validation set to tune the batch size. We search the set $\{8,16,32,64,100\}$ for the best batch size for varying number of labeled samples.

\subsubsection{$\Pi$-Model}

For the $\Pi$-Model, we minimize the loss $\loss_\textrm{Sup} + \eta_{\Pi}\loss_\textrm{Consistency}$ and observe that using the proposed training strategy by~\cite{laine2017temporal} leads to higher performance than training with our strategy but requires training for 150k iterations for both data sets: SVHN and CIFAR-10. Hence, we ramp-up the learning rate and the consistency regularization parameter $\eta_{\Pi}$ during the first 40k iterations and ramp-down the learning rate to zero and the parameter $\beta_1$ of the Adam optimizer to 0.5 during the last 25k iterations. We use their proposed ramp-up and ramp-down functions.
We use the validation set to select the best learning rate of the set $\{0.001, 0.003, 0.005\}$ and the maximal value of $\eta_{\Pi}$ after ramp-up from the set $\{1, 5, 10, 50, 100\}$. In contrast to~\cite{laine2017temporal}, we use the same data augmentation for all methods and omit adding Gaussian noise to the input samples.

\subsubsection{Mean Teacher}

Similar to the $\Pi$-Model, we observed better test performance of the Mean Teacher when using the training strategy proposed by~\cite{tarvainen2017meanteacher} to minimize the loss $\loss_\textrm{Sup} + \eta_{\textrm{MT}}\loss_\textrm{Consistency}$. Again, we train for 150k iterations for both data sets: SVHN and CIFAR-10. We ramp-up the learning rate and the consistency regularization parameter $\eta_{\textrm{MT}}$ during the first 40k iterations and ramp-down the learning rate to zero and the parameter $\beta_1$ of the Adam optimizer to 0.5 during the last 25k iterations. We use their proposed ramp-up and ramp-down functions, that are inspired by~\cite{laine2017temporal}.
Again, we use the validation set to select the best learning rate of the set $\{0.001, 0.003, 0.005\}$ and the maximal value of $\eta_{\textrm{MT}}$ after ramp-up from the set $\{1, 5, 10, 50, 100\}$. Again, we use for the mean-teacher the same data augmentation as for all other methods. We follow~\cite{tarvainen2017meanteacher} and update the teacher model parameters after each training step using exponential moving average with $\alpha = 0.999$.

\subsubsection{Virtual Adversarial Training}

\gls{vat} computes adversarial perturbations for labeled and unlabeled samples and requires consistent predictions even when the adversarial component is added to the input sample. Hence, the main hyper-parameter $\varepsilon$ of VAT controls the magnitude of the adversarial perturbation. Thus, we search for the best $\varepsilon$ using the validation set. We observe that the parameters $\varepsilon=3.5$ for SVHN and $\varepsilon=8$ proposed by~\cite{miyato2018vat} work best for all operating points for the corresponding data sets. \cite{miyato2018vat} studied the following loss function the loss $\loss_\textrm{Sup} + \eta_{\textrm{VAT}}\loss_\textrm{Consistency}$, but observed that tuning only $\varepsilon$ and keeping $\eta_{\textrm{VAT}} = 1$ leads to similar results as tuning both. Therefore, we use $\eta_{\textrm{VAT}} = 1$.

\subsubsection{Combinations with Distribution Alignment}

Combining distribution alignment and consistency regularization requires minimizing $\loss_\textrm{Sup} + \eta\loss_\textrm{Consistency} + \mu\loss_\textrm{Adv}$.
As soon as distribution alignment is involved independent of the other consistency based \gls{ssl} method, we use the training strategy introduced for distribution alignment. However, we use the method specific hyper-parameters and follow the corresponding tuning approach.

\subsection{Maximum Mean Discrepancy}

The \gls{mmd} score compares two different high dimensional distributions by measuring the difference between their means, see Eq.~\eqref{eq:mmd}.

\begin{align}\label{eq:mmd}
    \mathrm{MMD}_b^2(\Phi, \dataset{P},\dataset{Q}) = \sup_{\phi \in \Phi}\|\E_{\dataset{P}}[\phi(\vec{p})] - \E_{\dataset{Q}}[\phi(\vec{q})]\|_2^2 
    = \sup_{\phi \in \Phi}\biggl\|\frac{1}{|\dataset{P}|}\sum_{\vec{p}\in\dataset{P}}\phi(\vec{p}) - \frac{1}{|\dataset{Q}|}\sum_{\vec{q}\in\dataset{Q}}\phi(\vec{q})\biggl\|_2^2
\end{align}

Typically, we lack access to the true distributions and use instead two data sets $\dataset{P}$ and $\dataset{Q}$ that are sampled from the true distributions.
Hence, to avoid biased estimates, we use the unbiased empirical estimate of the square \gls{mmd} introduced by~\cite{gretton2012kernel}, that reads as

\begin{equation}
\begin{split}
    \mathrm{MMD}_{u}^{2}(\Phi, \dataset{P},\dataset{Q}) &= \frac{1}{|\dataset{P}|(|\dataset{P}|-1)}\sum_{i=1}^{|\dataset{P}|}\sum_{j\neq i}^{|\dataset{P}|}k(\vec{p}_i,\vec{p}_j) + \frac{1}{|\dataset{Q}|(|\dataset{Q}|-1)}\sum_{i=1}^{|\dataset{Q}|}\sum_{j \neq i}^{|\dataset{Q}|}k(\vec{q}_i,\vec{q}_j) \\
    &\quad - \frac{2}{|\dataset{P}||\dataset{Q}|}\sum_{i=1}^{|\dataset{P}|}\sum_{j=1}^{|\dataset{Q}|}k(\vec{p}_i,\vec{q}_j).
\end{split}
\end{equation}

Furthermore, using the introduced unbiased \gls{mmd} estimate only requires evaluating the value of a reproducing kernel Hilbert space instead of computing the full feature vector $\phi(\cdot)$ (kernel trick).

Therefore, the first metric that we use in the main paper uses the full feature distribution of the labeled and unlabeled samples. Then, the $\mathrm{MMD}^2$ score reads as 

\begin{equation}
    \mathrm{MMD}^2 \coloneqq \mathrm{MMD}_u^2(\Phi, \dataset{F}_l,\dataset{F}_{ul}),
\end{equation}

where $\dataset{F}_l = \{f(\img_l^i)\}_{i=1}^N$ and $\dataset{F}_{ul} = \{f(\img_{ul}^i)\}_{i=1}^M$ with the feature extractor $f$. We use the radial basis function kernel $\krbf(\vec{p}_i,\vec{q}_j) \coloneqq \exp{\brackets{-\|\vec{p}-\vec{q}\|_2^2/\gamma}}$
with the bandwidth $\gamma$ set to the median pairwise distance of the data $\dataset{F}_l$ and $\dataset{F}_{ul}$~\cite{gretton2012kernel,long2015learning} as the reproducing kernel Hilbert space.

\subsection{Derivation of the Approximate Upper Bound}

To derive the upper bound in Theorem 1, we parametrize the neural network by the weights $\W$ and bias components $\bias$. $f(\img)$ is the feature vector corresponding to the sample $\img$ extracted in front of the last fully connected layer. We can then introduce the output feature vector $z(x)$ by 

\begin{align}\label{eq:z_def}
    z(x) = \W f(x) + \bias
\end{align}

Furthermore, the $\lse(\vec{v}) = \log\brackets{\sum_i \exp{\brackets{v_i}}}$ function can be approximated by using the tangent line approximation of the logarithmic function $\log(a + b) \approx \log(a) + b/a$. The approximation holds in the vicinity of $a$ if $b$ is small. We then replace $a$ by a single value of the sum and $b$ by the sum of all other terms:

\begin{equation}\label{eq:lse-approx}
    \lse(\vec{v}) = \log(\exp(v_j) + \sum_{i\neq j}\exp(v_i)) \approx v_j + \frac{\sum_{i\neq j}\exp(v_i)}{\exp(v_j)}
\end{equation}

Now, when we assume that the component $v_j = \max_i v_i$ and that $\sum_{i\neq j}\exp(v_i) \ll \exp(v_j)$ we can simplify Eq.~\eqref{eq:lse-approx} even further such that

\begin{equation}\label{eq:lse-max-approx}
    \lse(\vec{v}) \approx \max_i v_i.
\end{equation}

Note, that as soon as the classifier produces unique predictions with high probability all terms of the sum except the maximal will be zero. Therefore, the tangent line approximation and the approximations in Eq.~\eqref{eq:lse-approx} and~\eqref{eq:lse-max-approx} are tight and hold with equality. 

Then, the upper bound for $\abs{\hat{R}(\DX_{ul}) - \hat{R}(\DX_l)}$ presented in Theorem 1 in the main paper can be derived as follows:

\begin{align}
    \abs{\hat{R}(\DX_{ul}) - \hat{R}(\DX_l)}
    &= \abs{\frac{1}{M}\sum_{i=1}^M\ell(h(\img_{ul}^i), \lab_{ul}^i) - \frac{1}{N}\sum_{i=1}^N\ell(h(\img_{l}^i), \lab_{l}^i)}\label{eq:full_line_1}
    \\
    &=\abs{\frac{1}{N}\sum_{i=1}^N\sum_{k=1}^K\delta_k(\lab_l^i)\log[\softmax(\z(\img_l^i))]_k - \frac{1}{M}\sum_{i=1}^M\sum_{k=1}^K\delta_k(\lab_{ul}^i)\log(\softmax[\z(\img_{ul}^i))]_k}\label{eq:full_line_2}
     \\
    &=\abs{\frac{1}{N}\sum_{i=1}^N\left[\sum_{k=1}^K\delta_k(\lab_l^i)\z(\img_l^i)_k -  \lse(\z(\img_l^i))\right] - \frac{1}{M}\sum_{i=1}^M\left[\sum_{k=1}^K\delta_k(\lab_{ul}^i)\z(\img_{ul}^i)_k - \lse(\z(\img_{ul}^i))\right]}\label{eq:full_line_3}
    \\
    &= \bigg|\sum_{k=1}^K \left[\frac{1}{N}\sum_{i=1}^N\delta_k(\lab_l^i)\z(\img_l^i)_k - \frac{1}{M}\sum_{i=1}^M\delta_k(\lab_{ul}^i)\z(\img_{ul}^i)_k\right] \\
    & \qquad + \left[\frac{1}{M}\sum_{i=1}^M\lse(\z(\img_{ul}^i)) - \frac{1}{N}\sum_{i=1}^N\lse(\z(\img_l^i))\right]\bigg|\label{eq:full_line_4}
    \\
    &\leq\sum_{k=1}^K\abs{\frac{1}{N}\sum_{i=1}^N\delta_k(\lab_l^i)\z(\img_l^i)_k - \frac{1}{M}\sum_{i=1}^M\delta_k(\lab_{ul}^i)\z(\img_{ul}^i)_k} \nonumber\\
    & \qquad + \abs{\frac{1}{M}\sum_{i=1}^M\lse(\z(\img_{ul}^i)) - \frac{1}{N}\sum_{i=1}^N\lse(\z(\img_l^i))}\label{eq:full_line_5}
    \\
    &\lessapprox\sum_{k=1}^K\abs{\frac{1}{N}\sum_{i=1}^N\delta_k(\lab_l^i)\z(\img_l^i)_k - \frac{1}{M}\sum_{i=1}^M\delta_k(\lab_{ul}^i)\z(\img_{ul}^i)_k} \nonumber\\
    & \qquad + \abs{\frac{1}{M}\sum_{i=1}^M\max(\z(\img_{ul}^i)) - \frac{1}{N}\sum_{i=1}^N\max(\z(\img_l^i))}\label{eq:full_line_6}
    \\
    &\lessapprox\sum_{k=1}^K\Biggl[\underbrace{\abs{\frac{1}{N}\sum_{i=1}^N\delta_k(\lab_l^i)\z(\img_l^i)_k - \frac{1}{M}\sum_{i=1}^M\delta_k(\lab_{ul}^i)\z(\img_{ul}^i)_k}}_{(\bigstar)}\nonumber\\ 
    & \qquad + \underbrace{\abs{\frac{1}{M}\sum_{i=1}^M\delta_k(h(\img_{ul}^i))\z(\img_{ul}^i)_k - \frac{1}{N}\sum_{i=1}^N\delta_k(h(\img_l^i))\z(\img_l^i)_k}}_{(\spadesuit)}\Biggl]\label{eq:full_line_7}
\end{align}

We plugin the cross entropy loss function where $\delta_p(q)$ is one if $p=q$ and zero otherwise, see Eq.~\eqref{eq:full_line_2}. Then, we reformulate in Eq.~\eqref{eq:full_line_3} the cross entropy function using the $\lse$ function. In Eq.~\eqref{eq:full_line_4} we reorder the terms and split the first term in a sum over all class labels. Then, the triangle inequality provides an upper bound of Eq.~\eqref{eq:full_line_4} in Eq.~\eqref{eq:full_line_5}. Finally, we use the $\lse$ approximation to obtain the approximate upper bound, see Eq.~\eqref{eq:full_line_6}. Then, in Eq.~\eqref{eq:full_line_7}, we use the fact that $\max z(x)$ and extracting the value $z(x)_k$ at position $k$ where $k$ corresponds to the most likely/predicted class are equivalent. Furthermore, we use the triangle inequality to split this last term in Eq.~\eqref{eq:full_line_6} in a sum over all class categories. Note, that both terms $(\bigstar)$ and $(\spadesuit)$ only differ by the argument of $\delta_k(\cdot)$. Therefore, we concentrate first on $(\bigstar)$ and derive from there $(\spadesuit)$.

The term $(\bigstar)$ can be further bounded:

\begin{align}
    (\bigstar) &= \abs{\frac{1}{N}\sum_{i=1}^N\delta_k(\lab_l^i)\left[\W f(\img_l^i) + \bias\right]_k - \frac{1}{M}\sum_{i=1}^M\delta_k(\lab_{ul}^i)\left[\W f(\img_{ul}^i) + \bias\right]_k}\label{eq:part_line_1}
    \\
    &= \abs{\frac{1}{N}\sum_{i=1}^N\delta_k(\lab_l^i)\left[\W f(\img_l^i)\right]_k - \frac{1}{M}\sum_{i=1}^M\delta_k(\lab_{ul}^i)\left[\W f(\img_{ul}^i)\right]_k + \frac{1}{N}\sum_{i=1}^N\delta_k(\lab_l^i)\bias_k - \frac{1}{M}\sum_{i=1}^M\delta_k(\lab_{ul}^i)\bias_k}\label{eq:part_line_2}
    \\
    &= \abs{\frac{1}{N}\sum_{i=1}^N\delta_k(\lab_l^i)\W_k f(\img_l^i) - \frac{1}{M}\sum_{i=1}^M\delta_k(\lab_{ul}^i)\W_k f(\img_{ul}^i) + \bias_k\left[\frac{1}{N}\sum_{i=1}^N\delta_k(\lab_l^i)- \frac{1}{M}\sum_{i=1}^M\delta_k(\lab_{ul}^i)\right]}\label{eq:part_line_3}
    \\
    &= \abs{\W_k\left[\frac{1}{N}\sum_{i=1}^N\delta_k(\lab_l^i)f(\img_l^i) - \frac{1}{M}\sum_{i=1}^M\delta_k(\lab_{ul}^i)f(\img_{ul}^i)\right] + \bias_k\left[\frac{1}{N}\sum_{i=1}^N\delta_k(\lab_l^i)- \frac{1}{M}\sum_{i=1}^M\delta_k(\lab_{ul}^i)\right]}\label{eq:part_line_4}
    \\
    &\leq\abs{\W_k\left[\frac{1}{N}\sum_{i=1}^N\delta_k(\lab_l^i)f(\img_l^i) - \frac{1}{M}\sum_{i=1}^M\delta_k(\lab_{ul}^i)f(\img_{ul}^i)\right]} + \abs{\bias_k\left[\frac{1}{N}\sum_{i=1}^N\delta_k(\lab_l^i)- \frac{1}{M}\sum_{i=1}^M\delta_k(\lab_{ul}^i)\right]}\label{eq:part_line_5}
    \\
    &\leq\abs{\W_k}\abs{\frac{1}{N}\sum_{i=1}^N\delta_k(\lab_l^i)f(\img_l^i) - \frac{1}{M}\sum_{i=1}^M\delta_k(\lab_{ul}^i)f(\img_{ul}^i)} + \abs{\bias_k}\abs{\frac{1}{N}\sum_{i=1}^N\delta_k(\lab_l^i)- \frac{1}{M}\sum_{i=1}^M\delta_k(\lab_{ul}^i)}\label{eq:part_line_6}
\end{align}

We use the definition of $\z(\img)$ in Eq.~\eqref{eq:z_def} to rearrange the terms into a sample dependent and independent part, see Eq.~\eqref{eq:part_line_2}. We observe that $\bias_k$ is independent of the data samples $\img$. This leads to Eq.~\eqref{eq:part_line_3}. where we reformulated $(\W f(\img))_k$ and got $\W_k f(\img)$ where $\W_k$ is the $k^{\text{\small th}}$ row vector of $\W$. In Eq.~\eqref{eq:part_line_4} we factorize $\W_k$ and $\bias_k$.
In Eq.~\eqref{eq:part_line_5} we used the triangle inequality to split the sum into two parts. In Eq.~\eqref{eq:part_line_6} we use the triangle inequality again to move $\W_k$ outside of the absolute term, and $|ab| = |a||b|$ where $a.b \in \mathbb{R}$ to pull the bias in front of the other absolute term.

The inner term ($\spadesuit$) is bounded the same way as $(\bigstar)$ and differs only by the arguments of $\delta_k(\cdot)$ either we use the true or the predicted labels to compute the average representation per class:

\begin{align}
    (\spadesuit) &\leq\abs{\W_k}\abs{\frac{1}{N}\sum_{i=1}^N\delta_k(h(\img_l^i))f(\img_l^i) - \frac{1}{M}\sum_{i=1}^M\delta_k(h(\img_{ul}^i))f(\img_{ul}^i)} + \abs{\bias_k}\abs{\frac{1}{N}\sum_{i=1}^N\delta_k(h(\img_l^i))- \frac{1}{M}\sum_{i=1}^M\delta_k(h(\img_{ul}^i))}
\end{align}

The sum of the two terms ($\bigstar$) and ($\spadesuit$) over all class categories $K$ leads to the upper bound of Theorem 1.

\begin{equation}
\begin{split}
    \sum_{k=1}^K\bracketss{(\bigstar) + (\spadesuit)} \leq \sum_{k=1}^K \Biggl[\abs{\W_k}&\abs{\frac{1}{N}\sum_{i=1}^N\delta_k(\lab_l^i)f(\img_l^i) - \frac{1}{M}\sum_{i=1}^M\delta_k(\lab_{ul}^i)f(\img_{ul}^i)}\\
    + \abs{\bias_k}&\abs{\frac{1}{N}\sum_{i=1}^N\delta_k(\lab_l^i)- \frac{1}{M}\sum_{i=1}^M\delta_k(\lab_{ul}^i)}\\
    + \abs{\W_k}&\abs{\frac{1}{N}\sum_{i=1}^N\delta_k(h(\img_l^i))f(\img_l^i) - \frac{1}{M}\sum_{i=1}^M\delta_k(h(\img_{ul}^i))f(\img_{ul}^i)}\\
    + \abs{\bias_k}&\abs{\frac{1}{N}\sum_{i=1}^N\delta_k(h(\img_l^i))- \frac{1}{M}\sum_{i=1}^M\delta_k(h(\img_{ul}^i))}\Biggl]
\end{split}
\end{equation}

\subsection{Additional Experiments}

\subsubsection{MMD vs. Test Error}

\begin{figure}[tb]
\centering 
\subfloat[
150 labels\newline\hspace*{1.5em}
$81.2 \pm 1.2$   \newline\hspace*{1.5em}
$36.8 \pm 3.1$
]{\includegraphics[width=0.2\textwidth, keepaspectratio]{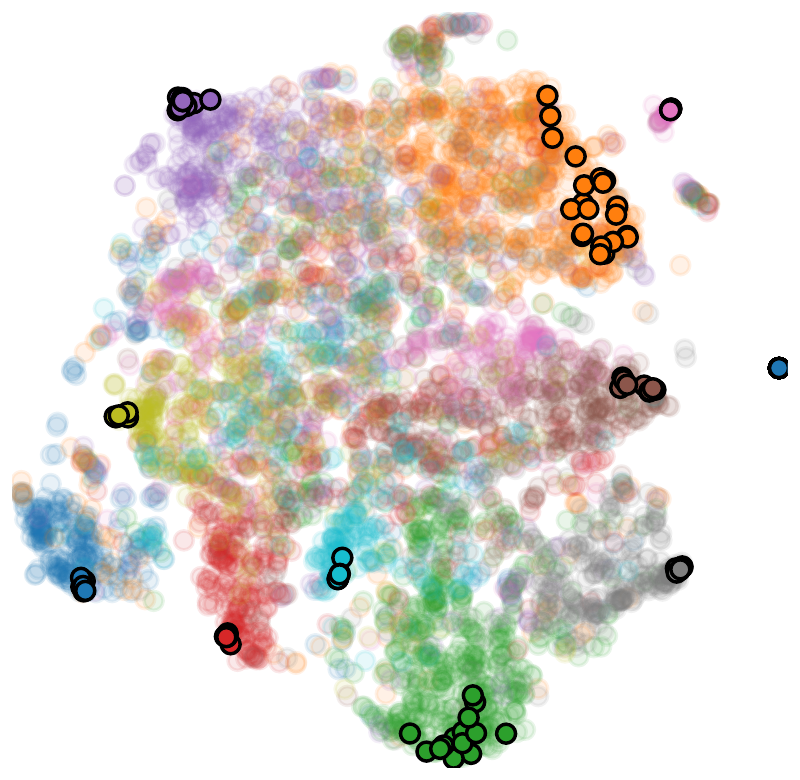}}
\subfloat[
200 labels\newline\hspace*{1.5em}
$27.3 \pm 3.2$\newline\hspace*{1.5em}
$29.8 \pm 1.4$
]{\includegraphics[width=0.2\textwidth, keepaspectratio]{tsne-svhn-fully-200-2-cropped.pdf}}
\subfloat[
250 labels\newline\hspace*{1.4em}
$36.4 \pm 1.4$\newline\hspace*{1.4em}
$26.0 \pm 1.4$
]{\includegraphics[width=0.2\textwidth, keepaspectratio]{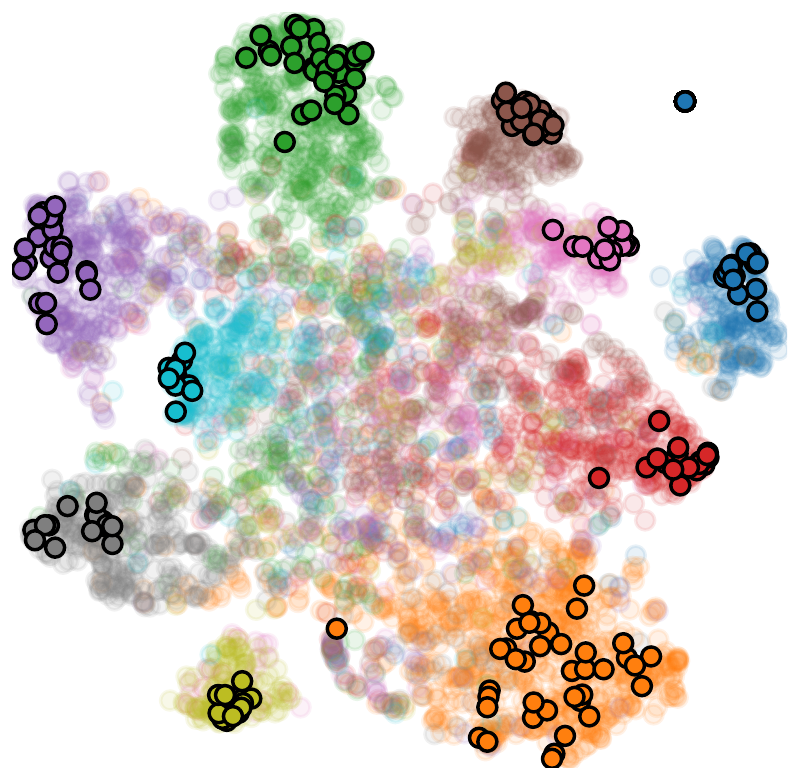}}
\subfloat[
500 labels\newline\hspace*{1.5em}
$9.9 \pm 1.6$\newline\hspace*{1.5em}
$15.4 \pm 1.1$
]{\includegraphics[width=0.2\textwidth,keepaspectratio]{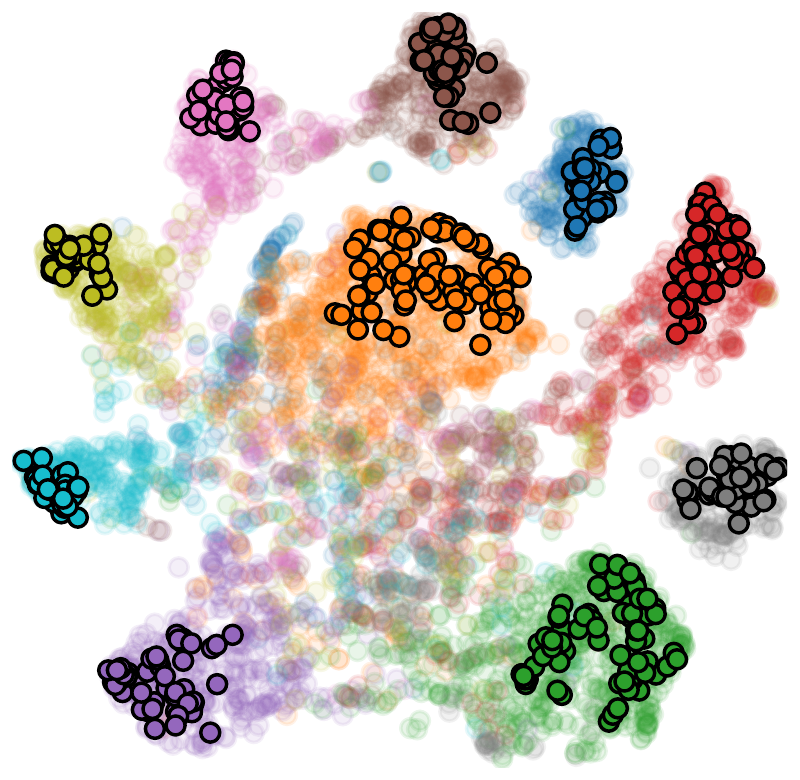}}
\subfloat[
1000 labels\newline\hspace*{1.5em}
$5.8 \pm 0.3$\newline\hspace*{1.5em}
$10.7 \pm 0.3$
]{\includegraphics[width=0.2\textwidth, keepaspectratio]{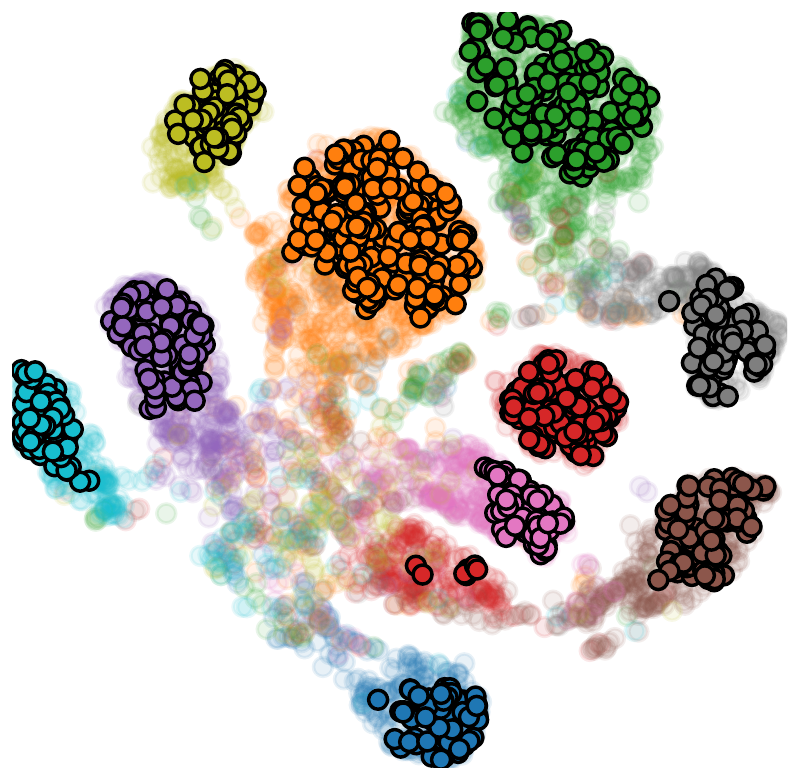}}

\caption{T-SNE~\cite{maaten2008visualizing} visualizations of the feature space for the supervised-only method using different amount of labeled samples. The dots with a black border correspond to labeled and all other to unlabeled samples, the colors identify the class correspondence. The models are trained with 200 labeled samples from SVHN~\cite{netzer2011svhn}. The plots show 5000 unlabeled samples. The 2nd row in the sub captions reports $\mathrm{MMD}^2$ ($\times 10^{-3}$) and the last row shows the test error (\%).}\label{fig:tsne-fully-all}

\end{figure}

\begin{figure}[t]
\centering 
\subfloat[
Supervised-only\newline\hspace*{1.5em}
$27.3 \pm 3.2$\newline\hspace*{1.5em}
$29.8 \pm 1.4$%
]{\includegraphics[width=0.25\textwidth, keepaspectratio]{tsne-svhn-fully-200-2-cropped.pdf}}
\subfloat[
$\Pi$-Model\newline\hspace*{1.5em}
$82.1 \pm 14.8$\newline\hspace*{1.5em}
$17.8 \pm 1.9$ 
]{\includegraphics[width=0.25\textwidth, keepaspectratio]{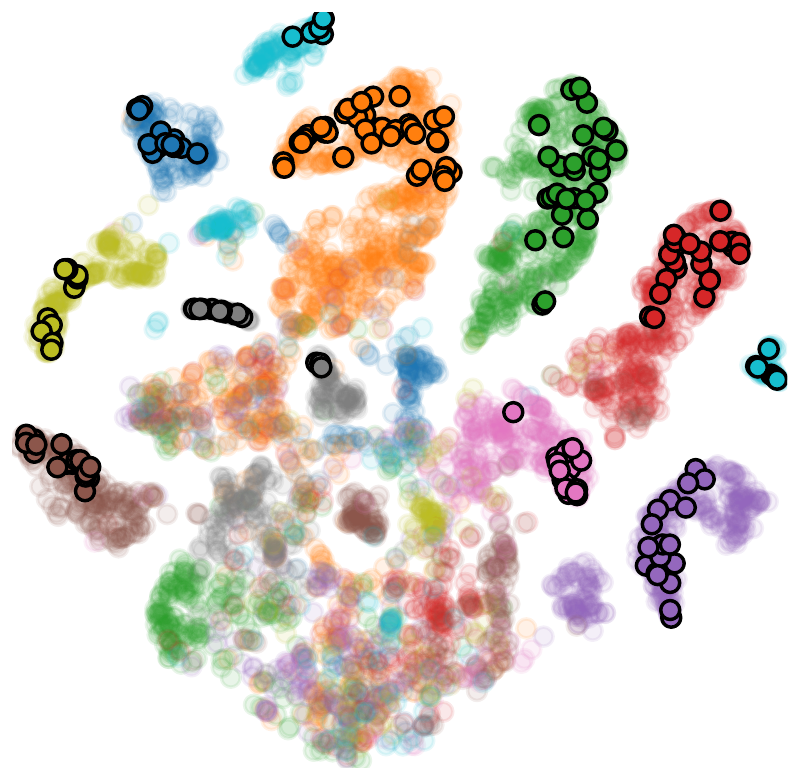}}
\subfloat[
Mean Teacher\newline\hspace*{1.5em}
$58.5 \pm 9.6$\newline\hspace*{1.5em}
$15.6 \pm 1.0$ 
]{\includegraphics[width=0.25\textwidth, keepaspectratio]{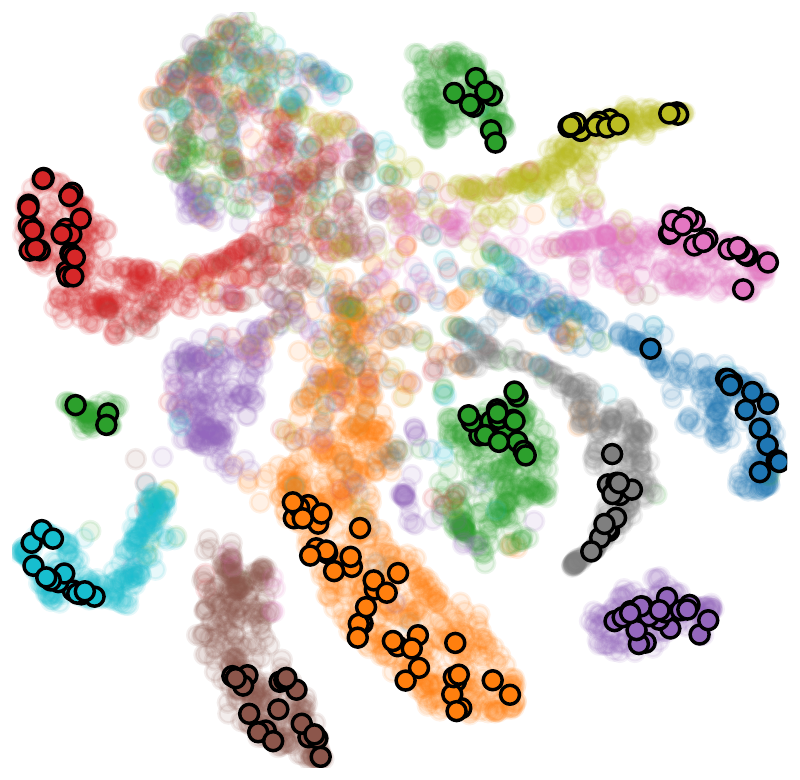}}
\subfloat[
VAT\hspace*{1.5em}\newline\hspace*{1.5em}
$13.7 \pm 1.9$\newline\hspace*{1.5em}
$13.4 \pm 1.6$ 
]{\includegraphics[width=0.25\textwidth,keepaspectratio]{tsne-svhn-vat-200-2-cropped.pdf}}
\\
\subfloat[
Alignment\newline\hspace*{1.5em}
$1.1 \pm 0.5$\newline\hspace*{1.5em}
$8.7 \pm 0.5$ 
]{\includegraphics[width=0.25\textwidth, keepaspectratio]{tsne-svhn-domain-200-2-cropped.pdf}}
\subfloat[
$\Pi$-Model with Alignment\newline\hspace*{1.5em}
$0.6 \pm 0.4$\newline\hspace*{1.5em}
$7.0 \pm 0.7$
]{\includegraphics[width=0.25\textwidth, keepaspectratio]{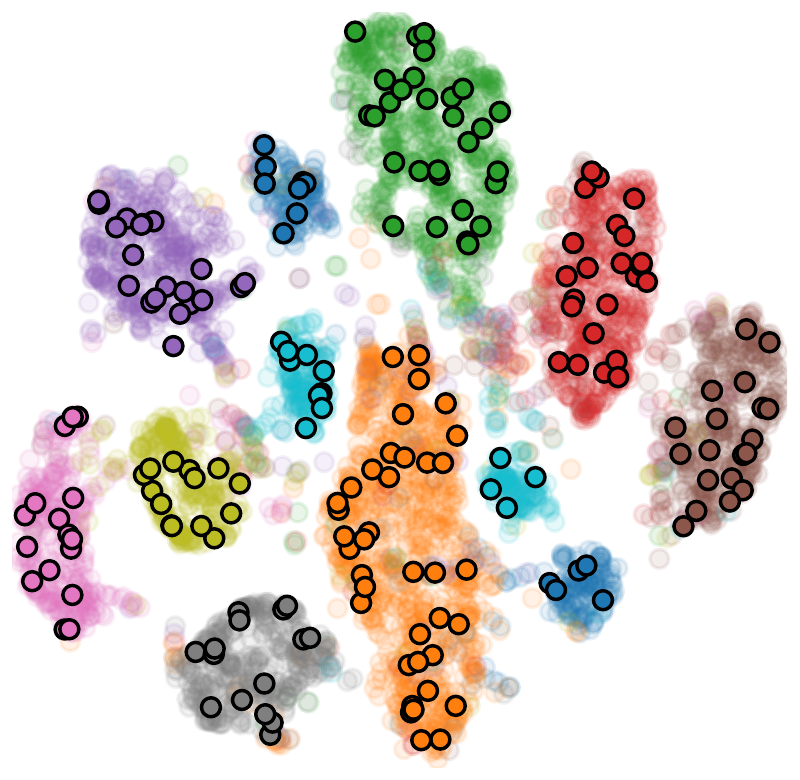}}
\subfloat[
Mean Teacher with Alignment\newline\hspace*{1.5em}
$0.82 \pm 1.4$\newline\hspace*{1.5em}
$6.9 \pm 0.7$
]{\includegraphics[width=0.25\textwidth, keepaspectratio]{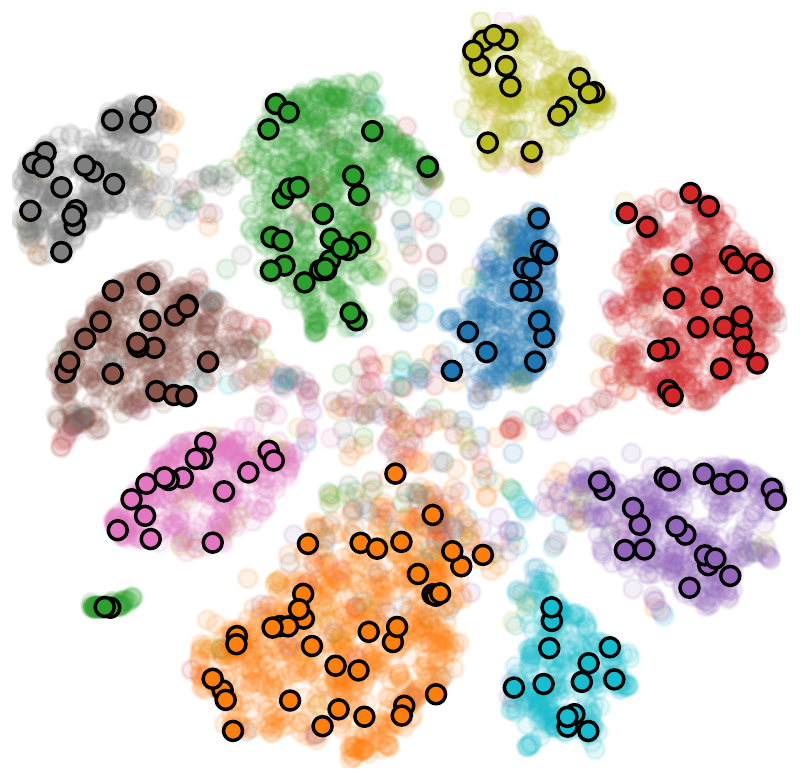}}
\subfloat[
VAT with Alignment\newline\hspace*{1.5em}
$2.3 \pm 0.4$\newline\hspace*{1.5em}
$4.9 \pm 0.1$ 
]{\includegraphics[width=0.25\textwidth,keepaspectratio]{tsne-svhn-domain-vat-200-2-cropped.pdf}}

\caption{T-SNE visualizations of the feature space for different \acrshort{ssl} training strategies without and with our distribution alignment method.  The dots with a black border correspond to labeled and all other to unlabeled samples, the colors identify the class correspondence. The models are trained with 200 labeled samples from SVHN~\cite{netzer2011svhn}. The plots show 200 labeled and 5000 unlabeled samples. The 2nd row in the sub captions reports $\mathrm{MMD}^2$ ($\times 10^{-3}$) and the last row shows the test error (\%).}\label{fig:tsne-improvement}

\end{figure}

Tab. \ref{tab:mmd} reports experimental results for the two considered metrics, of all considered \gls{ssl} methods in this paper and for both data sets at different operating points. These numbers backup the summarized view presented in Figs.~\ref{fig:tsne-ablation} and~\ref{fig:mmd-plots}. In addition, we show in Fig.~\ref{fig:tsne-fully-all} how the feature embedding changes when using more labeled samples to train the supervised-only method. Clearly, we observe that more annotations improve the distribution alignment.
Fig.~\ref{fig:tsne-improvement} shows a comparison of the feature embedding with and without distribution alignment for different \gls{ssl} methods. We conclude that the improvement obtained by adding distribution alignment translates to a tighter aligned feature embedding.

\subsubsection{Adversarial Distribution Alignment Improvements}

Tabs. \ref{tab:svhn-improvement} and \ref{tab:cifar-improvement} present all the operating points that were evaluated on the two data sets. In both cases, they allow to confirm the previously presented trend: Adversarial distribution alignment helps reducing the test errors in all cases and shines especially when very few labels are used.

\subsubsection{Experiments with Wide Residual Networks}

Additionally to the advantages presented earlier, we highlight the independence of our method to the proposed CNN architecture, by re-running the best performing experiments using a Wide ResNet~\cite{zagoruyko2016wide}. More specifically, we use WRN-28-2, that was recently proposed for \gls{ssl}~\cite{oliver2018realistic}. We use the same training strategy for the wide residual net as for the previously introduced network.

We also repeat the experiments without consistency and VAT~\cite{miyato2018vat} when using a wide residual net. The results (see Tab.~\ref{tab:svhn-resnet-improvement}) emphasize the even higher performance of our method with a wide residual network:
\begin{itemize}
    \item Adversarial distribution alignment reduces the test error for any operating point by more than 2.47.
    \item VAT with adversarial distribution alignment achieves a test error of only 4.83\% using 150 labels. In comparison, the fully-supervised model (requiring 73k labels), achieves an error of 3.38\%.
\end{itemize}

\begin{table*}[tb]
  \renewcommand{\arraystretch}{1.2}
  \newcommand{\head}[1]{\textnormal{\textbf{#1}}}
  \newcommand{\normal}[1]{\multicolumn{1}{c}{#1}}
  \newcommand{\na}{\emph{n.a.}}

  \colorlet{tableheadcolor}{gray!25} 
  \newcommand{\headcol}{\rowcolor{tableheadcolor}} %
  \colorlet{tablerowcolor}{gray!10} 
  \newcommand{\rowcol}{\rowcolor{tablerowcolor}} %
  \setlength{\aboverulesep}{0pt}
  \setlength{\belowrulesep}{0pt}

  \newcommand*{\rulefiller}{
    \arrayrulecolor{tableheadcolor}
    \specialrule{\heavyrulewidth}{0pt}{-\heavyrulewidth}
    \arrayrulecolor{black}}
  
    \centering
  \caption{Comparison between the two feature distribution alignment measures $\mathrm{MMD}^2$ $(\times10^{-3})$ and $\mathrm{mMMD}^2$ $(\times10^{-3})$ and the test error (\%) for different data sets, methods and operating points.}
  \label{tab:mmd}

\resizebox{1.\textwidth}{!}{%
  \begin{tabular}{ll*{8}{r}}
    \toprule
    \headcol                                                         &          & \multicolumn{3}{c}{SVHN}                                   & \multicolumn{3}{c}{CIFAR}\\
    \cmidrule(lr){3-5}\cmidrule(lr){6-8}
    \headcol                                                         &          & 200 labels          & 500 labels        & 1000 labels      & 250 labels         & 1000 labels       & 4000 labels   \\
    \headcol                                                         &          & 73257 images        & 73257 images      & 73257 images     & 50000 images       & 50000 images      & 50000 images \\
    \toprule
    \multirow{2}{*}{\makecell{\head{Supervised-}\\\head{only}}}      & MMD      & $27.34 \pm 3.20$    & $9.90 \pm 1.60$   & $5.79 \pm 0.31$  & $33.03 \pm 1.08$   & $5.28 \pm 0.30$   & $1.39 \pm 0.38$ \\
                                                                     & Test Err. & $29.82 \pm 1.37$    & $15.43 \pm 1.07$  & $10.74 \pm 0.28$ & $54.69 \pm 1.32$   & $35.79 \pm 1.45$  & $19.74 \pm 0.28$\\
    \midrule
    \multirow{2}{*}{\head{$\mathbf{\Pi}$ Model}}                     & MMD      & $82.08 \pm 14.77$   & $8.44 \pm 1.37$   & $10.62 \pm 1.33$ & $91.28 \pm 7.05$   & $38.05 \pm 2.03$  & $2.30 \pm 0.25$\\
                                                                     & Test Err. & $17.80 \pm 1.84$    & $8.81 \pm 0.28$   & $5.16  \pm 0.28$ & $53.68 \pm 1.50$   & $34.27 \pm 1.45$  & $15.98 \pm 0.16$ \\
    \midrule
    \multirow{2}{*}{\makecell{\head{Mean}\\\head{Teacher}}}          & MMD      & $58.49 \pm 9.63$    & $30.77 \pm 5.12$  & $19.64 \pm 3.80$ & $69.60 \pm 19.38$  & $21.16 \pm 2.53$  & $4.06 \pm 0.26$\\
                                                                     & Test Err. & $15.62 \pm 0.97$    & $6.28 \pm 0.11$   & $5.03 \pm 0.19$  & $53.09 \pm 1.60$   & $33.29 \pm 0.99$  & $15.35 \pm 0.18$\\
    \midrule
    \multirow{2}{*}{\head{VAT}}                                      & MMD      & $13.73 \pm 1.94$    & $5.44 \pm 0.13$   & $2.20 \pm 0.25$  & $27.91 \pm 1.45$   & $7.62 \pm 2.06$   & $1.19 \pm 0.20$\\
                                                                     & Test Err. & $13.41 \pm 1.55$    & $7.35 \pm 0.38$   & $5.91 \pm 0.37$  & $39.90 \pm 3.73$   & $23.08 \pm 2.87$  & $12.59 \pm 0.21$ \\
    \midrule
    \multirow{2}{*}{\makecell{\head{Adversarial}\\\head{Alignment}}}& MMD      & $1.13 \pm 0.49$     & $1.09 \pm 0.31$   & $0.66 \pm 0.23$  & $39.85 \pm 2.11$   & $0.80 \pm 0.15$   & $0.48 \pm 0.22$ \\
                                                                     & Test Err. & $8.72 \pm 0.53$     & $7.68 \pm 0.16$   & $7.54 \pm 0.26$  & $43.79 \pm 1.67$   & $25.17 \pm 0.60$  & $15.55 \pm 0.37$\\
    \bottomrule
  \end{tabular}
  }
\end{table*}

\begin{table*}[tb]
  \renewcommand{\arraystretch}{1.2}
  \newcommand{\head}[1]{\textnormal{\textbf{#1}}}
  \newcommand{\normal}[1]{\multicolumn{1}{c}{#1}}
  \newcommand{\na}{\emph{n.a.}}

  \colorlet{tableheadcolor}{gray!25} 
  \newcommand{\headcol}{\rowcolor{tableheadcolor}} %
  \colorlet{tablerowcolor}{gray!10} 
  \newcommand{\rowcol}{\rowcolor{tablerowcolor}} %
  \setlength{\aboverulesep}{0pt}
  \setlength{\belowrulesep}{0pt}

  \newcommand*{\rulefiller}{
    \arrayrulecolor{tableheadcolor}
    \specialrule{\heavyrulewidth}{0pt}{-\heavyrulewidth}
    \arrayrulecolor{black}}
  
    \centering
  \caption{Test errors (\%) for different \gls{ssl} methods with and without distribution alignment on SVHN. The fully-supervised model with 73k labeled training samples achieves a test error of $2.89 \pm 0.05$.}
  \label{tab:svhn-improvement}
  \begin{tabular}{lc*{5}{r}}
    \toprule
    \headcol                                     & $\loss_\textrm{Adv}$         & 150 labels       & 200 labels       & 250 labels       & 500 labels       & 1000 labels      \\
    \toprule
    \multirow{2}{*}{\head{No Consistency}}       & ---        & $36.81 \pm 3.08$ & $29.82 \pm 1.37$ & $26.04 \pm 1.43$ & $15.43 \pm 1.07$ & $10.74 \pm 0.28$ \\
                                                 & \checkmark & $9.09 \pm 1.08$  & $8.72 \pm 0.53$  & $8.43 \pm 0.13$  & $7.68 \pm 0.16$  & $7.54 \pm 0.26$ \\
    \midrule
    Reduction                                    &            & 27.72           & 21.10           & 16.61           & 7.75            & 3.20            \\
    \midrule
    \midrule
    \multirow{2}{*}{\head{$\mathbf{\Pi}$ Model}} & ---        & $31.23 \pm 1.93$ & $17.80 \pm 1.84$ & $11.98 \pm 0.54$ & $8.81 \pm 0.28$  & $6.12 \pm 0.24$ \\
                                                 & \checkmark & $8.62 \pm 0.36$  & $6.96 \pm 0.71$  & $6.54 \pm 0.19$  & $6.28 \pm 0.11$  & $5.16 \pm 0.28$ \\
    \midrule                                             
    Reduction                                    &            & 22.61            & 8.84            & 5.44            & 2.53            & 0.96            \\
    \midrule
    \midrule
    \multirow{2}{*}{\head{Mean Teacher}}         & ---        & $29.69 \pm 7.92$ & $15.62 \pm 0.97$ & $11.85 \pm 0.91$ & $7.07 \pm 0.41$ & $5.03 \pm 0.19$ \\
                                                 & \checkmark & $7.48 \pm 1.25$  & $6.93 \pm 0.70$  & $6.68 \pm 0.23$  & $5.67 \pm 0.11$ & $5.14 \pm 0.20$ \\
    \midrule
    Reduction                                    &            & 12.21           & 8.69            & 4.17            & 1.40           & -0.09            \\ 
    \midrule
    \midrule
    \multirow{2}{*}{\head{VAT}}                  & ---        & $18.30 \pm 4.02$ & $13.41 \pm 1.55$ & $10.59 \pm 0.98$ & $7.35 \pm 0.38$  & $5.91 \pm 0.37$ \\
                                                 & \checkmark & $8.93 \pm 3.24$  & $4.91 \pm 0.12$  & $4.71 \pm 0.12$  & $4.67 \pm 0.24$  & $4.93 \pm 0.07$ \\
    \midrule
    Reduction                                    &            & 9.63            & 8.50            & 5.88            & 2.68            & 0.98            \\
    \bottomrule
  \end{tabular}
\end{table*}

\begin{table*}[tb]
  \renewcommand{\arraystretch}{1.2}
  \newcommand{\head}[1]{\textnormal{\textbf{#1}}}
  \newcommand{\normal}[1]{\multicolumn{1}{c}{#1}}
  \newcommand{\na}{\emph{n.a.}}

  \colorlet{tableheadcolor}{gray!25} 
  \newcommand{\headcol}{\rowcolor{tableheadcolor}} %
  \colorlet{tablerowcolor}{gray!10} 
  \newcommand{\rowcol}{\rowcolor{tablerowcolor}} %
  \setlength{\aboverulesep}{0pt}
  \setlength{\belowrulesep}{0pt}

  \newcommand*{\rulefiller}{
    \arrayrulecolor{tableheadcolor}
    \specialrule{\heavyrulewidth}{0pt}{-\heavyrulewidth}
    \arrayrulecolor{black}}
  
    \centering
  \caption{Test errors (\%) for all \gls{ssl} methods with and without distribution alignment on CIFAR-10. The fully-supervised model with 50k labeled training samples achieves a test error of $6.87 \pm 0.14$}
  \label{tab:cifar-improvement}
  \begin{tabular}{lc*{5}{r}}
    \toprule
    \headcol                                     & $\loss_\textrm{Adv}$         & 250 labels       & 500 labels       & 1000 labels      & 2000 labels      & 4000 labels      \\
    \toprule
    \multirow{2}{*}{\head{No Consistency}}       & ---        & $54.69 \pm 1.32$ & $45.69 \pm 1.37$ & $35.79 \pm 1.45$ & $26.71 \pm 0.89$ & $19.74 \pm 0.28$ \\
                                                 & \checkmark & $43.79 \pm 1.67$ & $35.36 \pm 1.09$ & $25.17 \pm 0.60$ & $18.96 \pm 0.50$ & $15.55 \pm 0.37$ \\
    \midrule
    Reduction                                    &            & 10.90            & 10.34            & 10.62            & 7.75             & 4.19             \\
    \midrule
    \midrule
    \multirow{2}{*}{\head{$\mathbf{\Pi}$ Model}} & ---        & $53.68 \pm 1.50$ & $44.41 \pm 0.79$ & $34.27 \pm 1.45$ & $22.31 \pm 0.70$ & $15.98 \pm 0.16$ \\
                                                 & \checkmark & $43.42 \pm 2.03$ & $34.00 \pm 1.11$ & $23.40 \pm 0.49$ & $17.51 \pm 0.50$ & $13.64 \pm 0.29$ \\
    \midrule                                             
    Reduction                                    &            & 10.26            & 10.41            & 10.87            & 4.80             & 2.34            \\
    \midrule
    \midrule
    \multirow{2}{*}{\head{Mean Teacher}}         & ---        & $53.09 \pm 1.60$ & $44.47 \pm 1.13$ & $33.29 \pm 0.99$ & $21.40 \pm 0.44$ & $15.35 \pm 0.18$ \\
                                                 & \checkmark & $41.75 \pm 1.52$ & $33.27 \pm 0.82$ & $24.23 \pm 1.22$ & $17.37 \pm 0.62$ & $13.69 \pm 0.23$ \\
    \midrule
    Reduction                                    &            & 11.34            & 11.20            & 9.06             & 4.03             & 1.67             \\ 
    \midrule
    \midrule
    \multirow{2}{*}{\head{VAT}}                  & ---        & $39.90 \pm 3.73$ & $27.95 \pm 3.56$ & $23.08 \pm 2.87$ & $15.62 \pm 0.30$ & $12.59 \pm 0.21$ \\
                                                 & \checkmark & $29.44 \pm 2.00$ & $20.47 \pm 0.66$ & $16.35 \pm 0.36$ & $13.85 \pm 0.37$ & $11.54 \pm 0.16$ \\
    \midrule
    Reduction                                    &            & 10.46            & 7.48             & 6.73             & 1.78             & 1.05             \\
    \bottomrule
  \end{tabular}
\end{table*}

\begin{table*}[tb]
  \renewcommand{\arraystretch}{1.2}
  \newcommand{\head}[1]{\textnormal{\textbf{#1}}}
  \newcommand{\normal}[1]{\multicolumn{1}{c}{#1}}
  \newcommand{\na}{\emph{n.a.}}

  \colorlet{tableheadcolor}{gray!25} 
  \newcommand{\headcol}{\rowcolor{tableheadcolor}} %
  \colorlet{tablerowcolor}{gray!10} 
  \newcommand{\rowcol}{\rowcolor{tablerowcolor}} %
  \setlength{\aboverulesep}{0pt}
  \setlength{\belowrulesep}{0pt}

  \newcommand*{\rulefiller}{
    \arrayrulecolor{tableheadcolor}
    \specialrule{\heavyrulewidth}{0pt}{-\heavyrulewidth}
    \arrayrulecolor{black}}
  
    \centering
  \caption{Test errors (\%) of WRN-28-2 for all \gls{ssl} methods with and without distribution alignment on SVHN. The fully-supervised model with 72k labeled training samples achieves $3.38 \pm 0.05$.}
  \label{tab:svhn-resnet-improvement}
  \begin{tabular}{lc*{5}{r}}
    \toprule
    \headcol                                     & $\loss_\textrm{Adv}$         & 150 labels       & 200 labels       & 250 labels       & 500 labels       & 1000 labels      \\
    \toprule
    \multirow{2}{*}{\head{No Consistency}}       & ---        & $32.34 \pm 1.43$ & $27.86 \pm 1.30$ & $22.93 \pm 1.27$ & $15.39 \pm 0.63$ & $11.45 \pm 0.45$ \\
                                                 & \checkmark & $11.44 \pm 1.42$ & $8.94 \pm 0.42$  & $8.44 \pm 0.37$  & $8.05 \pm 0.32$  & $7.87 \pm 0.30$ \\
    \midrule
    Reduction                                    &            & 20.90            & 18.92            & 14.49            & 7.34             & 3.58            \\
    \midrule
    \midrule
    \multirow{2}{*}{\head{VAT}}                  & ---        & $16.41 \pm 1.62$ & $16.66 \pm 5.55$ & $11.64 \pm 2.65$ & $8.36 \pm 0.23$ & $6.67 \pm 0.38$ \\
                                                 & \checkmark & $4.83 \pm 0.21$  & $4.75 \pm 0.12$  & $4.80 \pm 0.21$  & $4.16 \pm 0.28$ & $4.20 \pm 0.14$ \\
    \midrule
    Reduction                                    &            & 11.58            & 11.91            & 6.84             & 4.20            & 2.47\\
    \bottomrule
  \end{tabular}
\end{table*}

\clearpage

\end{document}